\documentclass[twoside]{article}

\usepackage[accepted]{aistats2020}

\usepackage[round]{natbib}

\usepackage[toc,page]{appendix}

\usepackage{algpseudocode}
\usepackage{algorithm}
\usepackage{diagbox}
\usepackage{multicol}
\usepackage[colorlinks,linktoc=all]{hyperref}       %
\usepackage[usenames,dvipsnames]{xcolor}
\hypersetup{citecolor=Blue}
\hypersetup{linkcolor=Blue}
\hypersetup{urlcolor=Blue}
\usepackage{blindtext}
\usepackage{url}            %
\usepackage{booktabs}       %
\usepackage{amsfonts}       %
\usepackage{nicefrac}       %
\usepackage{amsthm, amsmath, amssymb}
\usepackage{graphicx, float, xcolor, subcaption, bm}
\usepackage{multirow}
\usepackage{makecell}
\usepackage{cleveref}
\usepackage{makecell}
\usepackage[normalem]{ulem}
\makeatletter
\newcommand*{\saved@uline}{}
\let\saved@uline\uline

\newcommand*{\mathuline}{%
	\mathpalette{\math@uline\saved@uline}%
}
\newcommand*{\math@uline}[3]{%
	\mbox{#1{$#2#3\m@th$}}%
}

\renewcommand*{\uline}{%
	\relax  
	\ifmmode
	\expandafter\mathuline
	\else
	\expandafter\saved@uline
	\fi
}
\makeatother

\theoremstyle{plain}

\theoremstyle{definition}

\newcommand{\cov}{\mathrm{Cov}}

\newcommand{\fin}{\bbf_{\|}} 
\newcommand{\fperp}{\bbf_{\perp}}

\newcommand{\vperp}{\bv_{\perp}}
\newcommand{\ppperp}{p_{\perp}}
\newcommand{\Kuu}{\bK_{\bu\bu}}
\newcommand{\Kuf}{\bK_{\bu\bbf}}
\newcommand{\Kfu}{\bK_{\bbf\bu}}
\newcommand{\Kff}{\bK_{\bbf\bbf}}
\newcommand{\Qff}{\bQ_{\bbf\bbf}}

\newcommand{\Kvv}{\bK_{\bv\bv}}
\newcommand{\Kvf}{\bK_{\bv\bbf}}

\newcommand{\Kvu}{\bK_{\bv\bu}}
\newcommand{\Kuv}{\bK_{\bu\bv}}
\newcommand{\Cff}{\bC_{\bbf\bbf}}
\newcommand{\Cvv}{\bC_{\bv\bv}}
\newcommand{\Cvf}{\bC_{\bv\bbf}}
\newcommand{\Cfv}{\bC_{\bbf\bv}}

\newcommand{\Lv}{\bL_\bv}
\newcommand{\Lu}{\bL_\bu}

\newcommand{\diag}{\text{diag}}

\newcommand{\bzero}{\textbf{0}}
\newcommand{\ba}{\textbf{a}}

\newcommand{\bbm}{\textbf{m}}

\newcommand{\bo}{\textbf{o}}

\newcommand{\bu}{\textbf{u}}
\newcommand{\bv}{\textbf{v}}
\newcommand{\bw}{\textbf{w}}
\newcommand{\bx}{\textbf{x}}
\newcommand{\by}{\textbf{y}}
\newcommand{\bz}{\textbf{z}}
\newcommand{\bbf}{\mathbf{f}}

\newcommand{\bA}{\textbf{A}}
\newcommand{\bB}{\textbf{B}}
\newcommand{\bC}{\textbf{C}}
\newcommand{\bD}{\textbf{D}}
\newcommand{\bE}{\textbf{E}}
\newcommand{\bF}{\textbf{F}}
\newcommand{\bG}{\textbf{G}}
\newcommand{\bH}{\textbf{H}}
\newcommand{\bI}{\textbf{I}}

\newcommand{\bK}{\textbf{K}}
\newcommand{\bL}{\textbf{L}}

\newcommand{\bO}{\textbf{O}}
\newcommand{\bP}{\textbf{P}}
\newcommand{\bQ}{\textbf{Q}}

\newcommand{\bS}{\textbf{S}}

\newcommand{\bX}{\textbf{X}}

\newcommand{\bZ}{\textbf{Z}}

\newcommand{\balpha}{\bm{\alpha}}

\newcommand{\bbeta}{\bm{\beta}}

\newcommand{\bsigma}{\bm{\sigma}}

\newcommand{\bmu}{\bm{\mu}}
\newcommand{\bSigma}{\bm{\Sigma}}

\newcommand{\KL}[2]{\mathrm{KL}\left[#1\|#2\right]}

\begin{document}

\twocolumn[

\aistatstitle{Sparse Orthogonal Variational Inference for Gaussian Processes}

\aistatsauthor{Jiaxin Shi \And Michalis K. Titsias \And Andriy Mnih}

\aistatsaddress{Tsinghua University \And DeepMind \And DeepMind} ]

\begin{abstract}
    We introduce a new interpretation of sparse variational approximations for Gaussian processes  using inducing points, which can lead to more scalable algorithms than previous methods. 
    It is based on decomposing a Gaussian process as a sum of two independent processes: one spanned by a finite basis of inducing points and the other capturing  the  remaining  variation.
    We show that this formulation recovers existing approximations and at the same time allows to obtain tighter lower bounds on the marginal likelihood and new stochastic  variational inference algorithms. We demonstrate the efficiency of these algorithms in several Gaussian process models ranging from standard regression to multi-class classification using (deep) convolutional Gaussian processes and report
    state-of-the-art results on CIFAR-10 among purely GP-based models.
\end{abstract}
\section{INTRODUCTION}

Gaussian processes (GP)~\citep{rasmussen2006gaussian} are nonparametric models for representing distributions over functions, which can be seen as a generalization of multivariate Gaussian distributions to infinite dimensions. The simplicity and elegance of these models has led to their wide adoption in uncertainty estimation for machine learning, including supervised learning~\citep{williams1996gaussian,williams1998bayesian}, sequential decision making~\citep{srinivas2010gaussian}, model-based planning~\citep{deisenroth2011pilco}, and unsupervised data analysis~\citep{lawrence2005probabilistic, damianou2016variational}.

Despite the successful application of these models, they suffer from $\mathcal{O}(N^3)$ computation and $\mathcal{O}(N^2)$ storage requirements given $N$ training data points, which has motivated a large body of research on sparse GP methods \citep{csato-opper-02,lawrence-seeger-herbrich-01, seeger03,candela-rasmussen-05,titsias2009variational,hensman2013gaussian,Buietal2017}.
GPs have also been unfavourably compared to deep learning models for lacking representation learning capabilities.

\begin{figure}[t]
    \centering
    \includegraphics[width=0.9\linewidth,trim=7.2cm 2.3cm 6.7cm 3cm,clip]{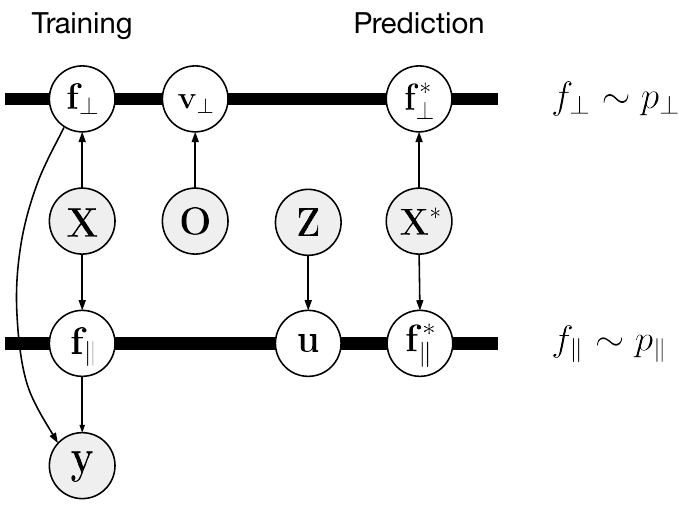}
    \caption{The graphical model of SOLVE-GP. The prior $f\sim \mathcal{GP}(0, k)$ is decomposed into two independent GPs (denoted by thick horizontal lines): $f_\|\sim p_\|$ and $f_\perp\sim p_\perp$. The variables connected by thick lines form a multivariate Gaussian. $\bX, \by$ denote the training data. $\bX^*$ are the test inputs. $\bbf_\| = f_\|(\bX)$, $\bbf_\perp = f_\perp(\bX)$. $\bu = f_\|(\bZ)$ denote the inducing variables in standard SVGP methods. SOLVE-GP introduces another set of inducing variables $\bv_\perp=f_\perp(\bO)$ to summarize $p_\perp$.}
    \label{fig:solve-gp}
     \vspace{-0.1in}
\end{figure}

Sparse variational GP (SVGP) methods~\citep{titsias2009variational,hensman2013gaussian,hensman2015scalable} based on variational learning of inducing points have shown promise in addressing these limitations. 
Such methods leave the prior distribution of the GP model unchanged and instead enforce sparse structures in the posterior approximation though variational inference. 
This gives $\mathcal{O}(M^2N + M^3)$ computation and $\mathcal{O}(MN + M^2)$ storage with $M$ inducing points. 
Moreover, they allow us to perform mini-batch training by sub-sampling data points. 
Successful application of SVGP allowed scalable GP models trained on billions of data 
points~\citep{salimbeni2017doubly}. 
These advances in inference methods have also led to more flexibility in model design. 
A recent convolutional GP model \citep{van2017convolutional} encodes translation invariance by summing over GPs that take image patches as inputs. 
The inducing points, which can be interpreted as image patches in this model, play a role similar to that of convolutional filters in neural networks. 
Their work showed that it is possible to implement representation learning in GP models. %
Further extensions of such models into deep hierarchies~\citep{blomqvist2018deep,dutordoir2019translation} significantly boosted the performance of GPs for natural images.

As these works suggest, currently the biggest challenge in this area still lies in scalable inference. 
The computational cost of the widely used SVGP methods scales cubically with the number of inducing points, making it difficult to improve the flexibility of posterior approximations~\citep{shi2019scalable}.
For example, state-of-the-art models like deep convolutional GPs use only 384 inducing points for inference in each layer to get a manageable running time~\citep{dutordoir2019translation}.
We introduce a new framework, called SOLVE-GP, which allows increasing the number of inducing points given a fixed computational budget.
It is based on decomposing the GP prior as the sum of a low-rank approximation using inducing points, and a full-rank residual process. 
We observe that the standard SVGP methods can be reinterpreted under such decomposition.
By introducing another set of inducing variables for the orthogonal complement, we can increase the number of inducing points at a much lower additional computational cost. 
With our method doubling the number of inducing points leads to a 2-fold increase in the cost of Cholesky decomposition, compared to the 8-fold increase for the original SVGP method.
We show that SOLVE-GP is equivalent to a structured covariance approximation for SVGP defined over the union of the two sets of inducing points. 
Interestingly, under such interpretation our work can be seen as a generalization of the recently proposed decoupled-inducing-points method~\citep{salimbeni2018orthogonally}. 
As the decoupled method often comes with a complex dual formation, our framework provides a simpler derivation and more intuitive understanding for it.

We conducted experiments on convolutional GPs and their deep variants. 
To the best of our knowledge, we are the first to train a purely GP-based model without any neural network components to achieve over $80\%$ test accuracy on CIFAR-10.
No data augmentation was used to obtain these results.
Besides classification, we also evaluated our method on a range of regression datasets that range in size from tens of thousands to millions of data points. 
Our results show that SOLVE-GP is often competitive with the more expensive SVGP counterpart that uses the same number of inducing points, and outperforms SVGP when given the same computational budget.

\section{BACKGROUND}
\label{sec:bg}

Here, we briefly review Gaussian processes and sparse variational GP methods.
A GP is an %
uncountable collection of random variables
indexed by a real-valued vector $\bx$ taking values in $\mathcal{X}\subset\mathbb{R}^d$,
of which any finite subset has a multivariate Gaussian distribution. 
A GP is defined by a mean function $m(\bx)=\mathbb{E}[f(\bx)]$ and a covariance function $k(\bx,\bx')=\cov[f(\bx),f(\bx')]$:
\begin{equation*}
    f \sim \mathcal{GP}(m(\bx), k(\bx,\bx')).
\end{equation*}
Let $\bX = [\bx_1, \bx_2, \dots, \bx_N]^\top \in \mathbb{R}^{N\times d}$ be (the matrix containing) the training data points and $\bbf = f(\bX)\in \mathbb{R}^N$ denote the corresponding function values. 
Similarly we denote the test data points by $\bX^*$ and their function values by $\bbf^*$. 
Assuming a zero mean function, the joint distribution over $\bbf,\bbf^*$ is given by:
\begin{equation*}
    p(\bbf, \bbf^*) := \mathcal{N}\left(\begin{array}{c} \bbf \\ \bbf^*\end{array} \middle|\;\mathbf{0} , \left[\begin{array}{cc}  \bK_{\bbf\bbf} & \bK_{\bbf*} \\ \mathbf{K}_{*\bbf} & \bK_{**} \end{array} \right] \right),
\end{equation*}
where $\bK_{\bbf\bbf}$ is an $N\times N$ kernel matrix with its $(i,j)$th entry as $k(\bx_i,\bx_j)$, and similarly $[\bK_{\bbf*}]_{ij} = k(\bx_i,\bx^*_j)$, $[\bK_{**}]_{ij} = k(\bx^*_i,\bx^*_j)$.
In practice we often observe the training function values through some noisy measurements $\by$, generated by the likelihood function $p(\by|\bbf)$. 
For regression, the likelihood usually models independent Gaussian observation noise:
$
y_n = f_n + \epsilon_n,\;\epsilon_n \sim \mathcal{N}(0, \sigma^2).
$
In this situation the exact posterior distribution $p(\bbf^*|\by)$ can be computed in closed form:
\begin{align} \label{eq:exact-gp}
\bbf^*|\by \sim \mathcal{N}(&\bK_{*\bbf}(\bK_{\bbf\bbf} + \sigma^2\bI)^{-1}\by, \notag \\
& \bK_{**} - \bK_{*\bbf}(\bK_{\bbf\bbf} + \sigma^2\bI)^{-1}\bK_{\bbf*}).
\end{align}
As seen from Eq.~\eqref{eq:exact-gp}, exact prediction involves %
the inverse of matrix $\bK_{\bbf\bbf} + \sigma^2\bI$, which requires $\mathcal{O}(N^3)$ computation. 
For large datasets, we need to avoid the cubic complexity by resorting to approximations.

Inducing points have played a central role in previous works on scalable GP inference. 
The general idea is to summarize $\bbf$ with a small number of variables $\bu = f(\bZ)$, where $\bZ = [\bz_1, \dots, \bz_M]^\top\in \mathbb{R}^{M\times d}$ is a set of parameters, called inducing points, in the input space. 
The augmented joint distribution over $\bu,\bbf,\bbf^*$ is $p(\bbf,\bbf^*|\bu)p(\bu)$, where $p(\bu) = \mathcal{N}(\bzero, \Kuu)$ and $\Kuu$ denotes the kernel matrix of inducing points with the $(i,j)$th entry corresponding to $k(\bz_i,\bz_j)$.
There is a long history of developing sparse approximations for GPs by making different independence assumptions for the conditional distribution $p(\bbf,\bbf^*|\bu)$ to reduce the computational cost~\citep{quinonero2005unifying}. 
However, these methods made modifications to the GP prior and tended to suffer from degeneracy and overfitting problems.

Sparse variational GP methods (SVGP), first proposed in \citet{titsias2009variational} and later extended for mini-batch training and non-conjugate likelihoods~\citep{hensman2013gaussian,hensman2015scalable}, provide an elegant solution to these problems. 
By reformulating the posterior inference problem as variational inference and restricting the variational distribution to be $q(\bbf,\bbf^*,\bu) := q(\bu)p(\bbf,\bbf^*|\bu)$, the variational lower bound for minimizing $\KL{q(\bbf,\bbf^*,\bu)}{p(\bbf,\bbf^*,\bu|\by)}$  simplifies to: 
\begin{equation} \label{eq:svgp}
\sum_{n=1}^N\mathbb{E}_{q(\bu)p(f_n|\bu)}\left[\log p(y_n|f_n)\right] - \KL{q(\bu)}{p(\bu)}.
\end{equation}
For GP regression the bound has a collapsed form obtained by solving for the optimal $q(\bu)$ and plugging it into~\eqref{eq:svgp}~\citep{titsias2009variational}: 
\begin{align}
\log \mathcal{N}(\by | \bzero, \bQ_{\bbf\bbf} + \sigma^2 \bI) 
- \frac{1}{ 2 \sigma^2} 
\text{tr}\left(  \bK_{\bbf\bbf} - \bQ_{\bbf\bbf} \right),
\label{eq:referencebound}
\end{align}
where $\bQ_{\bbf\bbf} = \bK_{\bbf\bu}\bK_{\bu\bu}^{-1}\bK_{\bu\bbf}$.
Computing this objective requires $\mathcal{O}(M^2N + M^3)$ operations, in contrast to the $\mathcal{O}(N^3)$ complexity of exact inference. The inducing points $\bZ$ can be learned as variational parameters by maximizing the lower bound.
More generally, if we do not collapse $q(\bu)$ and let $q(\bu) = \mathcal{N}(\bbm_\bu,\bS_\bu)$, where $\bbm_\bu, \bS_\bu$ are trainable parameters, we %
can use the uncollapsed bound for
mini-batch training and non-Gaussian 
likelihoods~\citep{hensman2013gaussian,hensman2015scalable}.

\section{SOLVE-GP}

Despite the success of SVGP methods, their $\mathcal{O}(M^3)$ complexity makes it difficult for the flexibility of posterior approximation to grow with the dataset size.
We present a new framework called \emph{Sparse OrthogonaL Variational infErence for Gaussian Processes}~(SOLVE-GP), 
which allows the use of an additional set of inducing points at a lower computational cost than the standard SVGP methods.  

\subsection{Reinterpreting SVGP}
\label{sec:repara}

We start by reinterpreting SVGP methods using a simple reparameterization, which will then lead us to possible ways of improving the approximation.
First we notice that the covariance of the conditional distribution $p(\bbf|\bu) = \mathcal{N}(\Kfu\Kuu^{-1}\bu, \Kff - \Qff)$ does not depend on $\bu$.\footnote{Note that kernel matrices like $\bK_{\bu\bu}$ depend on $\bZ$ instead of $\bu$; the subscript only indicates that this is the covariance matrix of $\bu$.}
Therefore, samples from $p(\bbf|\bu)$ can be reparameterized as
\begin{align} \label{eq:cond-repara}
    \fperp &\sim p_{\perp}(\fperp) := \mathcal{N}(\bzero, \Kff - \Qff), \notag \\
    \bbf &= \fperp + \Kfu\Kuu^{-1}\bu.
\end{align}
The reason for denoting the zero-mean component as $\fperp$ shall become clear later. 
Now we can reparameterize the augmented prior distribution $p(\bbf,\bu)$ as
\begin{align} \label{eq:prior-repara}
\bu \sim p(\bu),\quad\fperp \sim p_{\perp}(\fperp),\quad\bbf = \bK_{\bbf\bu}\bK_{\bu\bu}^{-1}\bu + \fperp,
\end{align}
and the joint distribution of the GP model becomes
\begin{multline}
p(\by,\bu,\fperp) = p(\by|\fperp + \bK_{\bbf\bu}\bK_{\bu\bu}^{-1}\bu)p(\bu) p_\perp(\fperp).
\label{eq:joint1}
\end{multline}
Posterior inference for $\bbf$ in the original model then turns %
into inference for $\bu$ and $\fperp$.
If we approximate the above GP model by considering a factorised 
approximation $q(\bu) p_\perp(\fperp)$, where $q(\bu)$ is a variational 
distribution and $p_\perp(\fperp)$ is the prior distribution of 
$\fperp$ that appears also in Eq.~\eqref{eq:joint1}, we arrive at the standard SVGP method. 
To see this, note that minimizing $\KL{q(\bu)p_{\perp}(\fperp)}{p(\bu, \fperp|\by)}$ is equivalent to maximizing the variational lower bound 
\begin{equation*}
\mathbb{E}_{q(\bu)p_\perp(\fperp)}\log p(\by|\fperp + \bK_{\bbf\bu}\bK_{\bu\bu}^{-1}\bu) - \KL{q(\bu)}{p(\bu)},
\end{equation*}
which is the SVGP objective (Eq.~\eqref{eq:svgp}) using the reparameterization in Eq.~\eqref{eq:cond-repara}.

Under this interpretation of the standard SVGP method, it becomes clear that we can modify the form of the variational distribution $q(\bu) p_\perp(\fperp)$ to improve the accuracy of the posterior approximation.
There are two natural options: (i) keep 
$p_\perp(\fperp)$ as part of the approximation and alter $q(\bu)$ so that it will have 
some dependence on $\fperp$, and (ii) keep $q(\bu)$ independent from $\fperp$, and 
replace $p_\perp(\fperp)$ with a more structured variational distribution 
$q(\fperp)$. 
While both options lead to new bounds and more accurate approximations than the standard method, we will defer the discussion of (i) to \cref{app:other-bounds} and focus on (ii) because it is amenable to large-scale training, as we will show next. 

\subsection{Orthogonal Decomposition}
\label{sec:orth}

As suggested in \cref{sec:repara}, we consider improving the variational distribution for $\fperp$.
However, the complexity of inferring $\fperp$ 
is the same as for $\bbf$ and thus cubic.
Resolving the problem requires a better understanding of the reparameterization we used in \cref{sec:repara}.

The key observation here is that the reparameterization 
in Eq.~\eqref{eq:prior-repara} corresponds to an orthogonal decomposition in the function space. 
For simplicity, we first derive such decomposition in the Reproducing Kernel Hilbert Space~(RKHS) induced by $k$, and then generalize the result to the GP sample space.
The RKHS with kernel $k$ is the closure of the space $\{\sum_{i=1}^\ell c_ik(\bx'_i,\cdot),\; c_i\in\mathbb{R}, \ell\in \mathbb{N}^+, \bx'_i \in \mathcal{X}\}$, with the inner product defined as $\langle f, k(\bx, \cdot)\rangle_{\mathcal{H}} = f(\bx),\; \forall f\in \mathcal{H}$.
Let $V$ denote the linear span of the kernel basis functions indexed by the inducing points:
$V := \{\sum_{j=1}^M \alpha_j k(\bz_j, \cdot),\; \balpha = [\alpha_1,\dots, \alpha_M]^\top \in \mathbb{R}^M\}.$
For any function $f\in \mathcal{H}$, we can decompose it~\citep{cheng2016incremental} as
\begin{align*}
f = f_\| + f_\perp,\quad f_\| \in V\text{ and }f_\perp \perp V,
\end{align*}
Assuming $f_\| = \sum_{j=1}^M \alpha_j' k(\bz_j, \cdot)$, then %
we can solve for the coefficients~(details in \cref{app:inner-product}): $\balpha' = k(\bZ,\bZ)^{-1}f(\bZ)$, where $k(\bZ, \bZ)$ denotes the kernel matrix of $\bZ$.
Therefore,
\begin{equation}
    \label{eq:func-orth-decomp}
    f_\|(\bx) = k(\bx, \bZ)k(\bZ,\bZ)^{-1}f(\bZ),\quad f_\perp = f - f_\|.
\end{equation}
Here $k(\bx, \bZ) := [k(\bz_1, \bx), \dots, k(\bz_M, \bx)]$. 
Although Eq.~\eqref{eq:func-orth-decomp} is derived by assuming $f\in\mathcal{H}$, %
it motivates us to study the same decomposition for $f\sim \mathcal{GP}(0, k)$.
Then $f_\|$ becomes $k(\cdot, \bZ)\Kuu^{-1}\bu$. 
Interestingly, we can verify that this is a sample from a GP with a zero mean function and covariance function $\cov[f_\|(\bx), f_\|(\bx')] = k(\bx,\bZ)\Kuu^{-1}k(\bZ,\bx')$. %
Similarly we can show that $f_\perp$ is a sample from another GP and we denote these two independent GPs as $p_{\|}$ and $\ppperp$~\citep{hensman2017variational}:
\begin{align*}
    f_\| \sim p_{\|} & \equiv \mathcal{GP}(0, k(\bx,\bZ)\Kuu^{-1}k(\bZ,\bx')), \\
    f_{\perp} \sim p_\perp &\equiv \mathcal{GP}(0, k(\bx,\bx') - k(\bx,\bZ)\Kuu^{-1}k(\bZ,\bx')).
\end{align*}
Marginalizing out the GPs at the training points $\bX$, it is easy to show that
\begin{align*}
\fin &= f_\|(\bX) = \bK_{\bbf\bu}\bK_{\bu\bu}^{-1}\bu \sim \mathcal{N}(\bzero, \Kfu\Kuu^{-1}\Kuf), \\
\fperp &= f_\perp(\bX) \sim \mathcal{N}(\bzero, \bK_{\bbf\bbf} - \Kfu\Kuu^{-1}\Kuf).
\end{align*}
This is exactly the decomposition we used in \cref{sec:repara}, and the meaning of $\fperp$  becomes clear.

\subsection{SOLVE-GP Lower Bound}
\label{sec:solvegp-lb}

The decomposition described in the previous section gives new insights for improving the variational distribution for $\fperp$. 
Specifically, we can introduce a second set of inducing variables $\vperp := f_{\perp}(\bO)$ to approximate $p_\perp$, as illustrated in Fig.~\ref{fig:solve-gp}. 
We call this second set $\bO = [\bo_1, \dots, \bo_{M_2}]^\top\in \mathbb{R}^{M_2\times d}$ the  \emph{orthogonal} inducing points. 
The joint model distribution is then
\begin{equation*}
p(\by|\fperp + \bK_{\bbf\bu}\bK_{\bu\bu}^{-1}\bu)p(\bu)p_\perp(\fperp|\vperp)p_\perp(\vperp).
\end{equation*}
First notice that the standard SVGP methods correspond to using the variational distribution $q(\bu)  p_\perp (\vperp) p_\perp(\fperp|\vperp)$.
To obtain better approximations we can replace the prior factor
$p_\perp (\vperp)$ with a tunable variational factor $q(\vperp) := \mathcal{N}(\bbm_\bv, \bS_\bv)$:
\begin{equation*}
q(\bu,\fperp,\vperp) = q(\bu)q(\vperp)p_\perp(\fperp|\vperp).
\end{equation*}
This gives the SOLVE-GP variational lower bound:
\begin{multline} \label{eq:sovgp}
\mathbb{E}_{q(\bu)q_\perp(\fperp)}\left[\log p(\by|\fperp + \bK_{\bbf\bu}\bK_{\bu\bu}^{-1}\bu)\right]  \\ -\KL{q(\bu)}{p(\bu)} - \KL{q(\vperp)}{p_\perp(\vperp)},
\end{multline}
where 
$q_{\perp}(\cdot) := \int p_\perp(\cdot|\vperp)q(\vperp) d\vperp$ is the variational predictive distribution for $p_\perp$. 
Simple computations show that $q_\perp(\fperp) = \mathcal{N}(\Cfv\Cvv^{-1}\bbm_{\bv}, \bS_{\fperp})$, where $\bS_{\fperp} = \Cff + \Cfv\Cvv^{-1}(\bS_\bv - \Cvv)\Cvv^{-1}\Cvf$. 
Here $\bC_{\bbf\bbf} := \bK_{\bbf\bbf}  - \bQ_{\bbf\bbf}$ is the covariance matrix of $p_\perp$ on the training inputs and similarly for the other matrices.
Because the likelihood factorizes given $\bbf$ (i.e., $\fperp +\Kfu\Kuu^{-1}\bu$), the first term of Eq.~\eqref{eq:sovgp} simplifies to $\sum_{n=1}^N \mathbb{E}_{q(\bu)q(f_\perp(\bx_n))}[\log p(y_n|f_\perp(\bx_n) + k(\bx_n, \bZ)\Kuu^{-1}\bu)]$.
Therefore, we only need to compute marginals of $q_\perp(\fperp)$ at individual data points. 
In the general setting, %
the SOLVE-GP lower bound can be maximized in $\mathcal{O}(N\bar{M}^2 + \bar{M}^3)$ time per gradient update, where $\bar{M}=\max(M,M_2)$. 
In mini-batch training $N$ is replaced by the batch size.
The predictive density at test data points 
can be found in \cref{app:pred}.

To intuitively understand the improvement over the standard SVGP methods, we derive a collapsed bound for GP regression using \eqref{eq:sovgp} and compare it to the \citet{titsias2009variational} bound.
Plugging in the optimal $q(\bu)$, and simplifying (see \cref{app:collapsed-solvegp}), gives the bound
\begin{multline}
\log \mathcal{N}(\by | \bC_{\bbf \bv} \bC_{\bv \bv}^{-1} \bbm_\bv,  \bQ_{\bbf\bbf} + \sigma^2 \bI)
-\frac{1}{2\sigma^2}\mathrm{tr}(\bS_{\fperp}) \\
- \KL{\mathcal{N}(\bbm_\bv,\bS_\bv)}{\mathcal{N}(\bzero,\bC_{\bv\bv})}.
\label{eq:collapsed2}
\end{multline}
With an appropriate choice of $q(\vperp)$
this bound can be 
tighter than the \citet{titsias2009variational} bound. 
For example, notice that when $q(\vperp)$ 
is equal to the prior $p_\perp (\vperp)$, i.e., $\bbm_\bv = {\bf 0}$ and $\bS_\bv = \bC_{\bv \bv}$, the bound in 
\eqref{eq:collapsed2} reduces to the one in \eqref{eq:referencebound}. 
Another 
interesting special case arises when the variational distribution has the same covariance matrix as the prior~(i.e., $\bS_\bv = \bC_{\bv \bv}$), while the mean $\bbm_\bv$ is learnable. Then the bound 
becomes 
\begin{multline}
\log \mathcal{N}(\by | \bC_{\bbf \bv} \bC_{\bv \bv}^{-1} \bbm_\bv,  \bQ_{\bbf\bbf} + \sigma^2 \bI) \\
- \frac{1}{2\sigma^2} 
\text{tr}\left( \bK_{\bbf \bbf}  -  \bQ_{\bbf \bbf} \right) - \frac{1}{2}\bbm_\bv^\top\bC_{\bv\bv}^{-1}\bbm_\bv.
\label{eq:collapsed3}
\end{multline}
Here we see that the second set of inducing variables 
$\vperp$ mostly determines the mean prediction over $\by$, which is zero in  the \citet{titsias2009variational} bound~(Eq.~\eqref{eq:referencebound}).

Our method introduces another set of inducing points to improve the variational approximation.
One natural question to ask is, how does this compare to the standard SVGP algorithm with the inducing points chosen to be union of the two sets?
We answer it as follows: 1) Given the same number of inducing points, SOLVE-GP is more computationally efficient than the standard SVGP method; 2) SOLVE-GP can be interpreted as using a structured covariance in the variational approximation for SVGP.

\paragraph{Computational Benefits.}
For a quick comparison, we analyze the cost of the Cholesky decomposition in both methods.
We assume the time complexity of decomposing an $M\times M$ matrix is $cM^3$, where $c$ is constant w.r.t.~$M$.
For SOLVE-GP, to compute the inverse and the determinant of $\Kuu$ and $\Cvv$, we need the Cholesky factors of them, which cost $c(M^3 + M_2^3)$.
For SVGP with $M$ inducing points, we need the Cholesky factor of $\Kuu$, which costs $cM^3$.
Adding another $M$ inducing points in SVGP leads to an 8-fold increase~(i.e., from $cM^3$ to $8cM^3$) in the cost of the Cholesky decomposition, compared to the 2-fold increase if we switch to SOLVE-GP with $M_2=M$ orthogonal inducing points.
A more rigorous analysis is given in \cref{app:pred}, where we enumerate all the cubic-cost operations needed when we compute the bound.

\paragraph{Structured Covariance.}
We can express our variational approximation w.r.t.\ the original GP. Let $\bv = f(\bO)$ denote the function outputs at the orthogonal inducing points. 
We then have the following relationship between $\bu,\bv$ and $\bu,\bv_\perp$:
\begin{equation*}
\begin{bmatrix}\bu \\ \bv\end{bmatrix} = \begin{bmatrix}\bI & \bzero \\ \bK_{\bv\bu}\bK_{\bu\bu}^{-1} & \bI\end{bmatrix}
\begin{bmatrix}\bu \\ \vperp\end{bmatrix}.
\end{equation*}
Therefore, the joint variational distribution over $\bu$ and $\bv$ that corresponds to the factorized $q(\bu)q(\vperp)$ is also Gaussian. 
By change-of-variable we can express it as $q(\bu,\bv) = \mathcal{N}(\bbm_{\bu,\bv},\bS_{\bu,\bv})$, where
$\bbm_{\bu,\bv} = \begin{bmatrix}
\bbm_\bu, \bbm_\bv + \bK_{\bv\bu}\bK_{\bu\bu}^{-1}\bbm_\bu
\end{bmatrix}^\top$
and %
\begin{equation*}
\bS_{\bu,\bv} = \begin{bmatrix}
\bS_\bu \!&\! \bS_\bu\bK_{\bu\bu}^{-1}\bK_{\bu\bv} \\
\bK_{\bv\bu}\bK_{\bu\bu}^{-1}\bS_\bu \!&\! \bS_\bv + \bK_{\bv\bu}\bK_{\bu\bu}^{-1}\bS_\bu\bK_{\bu\bu}^{-1}\bK_{\bu\bv}
\end{bmatrix}.
\end{equation*}
From $\bS_{\bu,\bv}$ we can see that our approach is different from making the mean-field assumption $q(\bu,\bv) = q(\bu)q(\bv)$, instead it captures the covariance between $\bu,\bv$ through a structured parameterization.

\section{EXTENSIONS}
\label{sec:ext}

One direct extension of SOLVE-GP involves using more than two sets of inducing points by repeatedly applying the decomposition.
However, this adds more complexity to the implementation.
Below we show that the SOLVE-GP framework can be easily extended to different GP models where the standard SVGP method applies.

\paragraph{Inter-domain and Convolutional GPs.}
Similar to SVGP methods, SOLVE-GP can deal with inter-domain inducing points~\citep{lazaro2009inter} which lie in a different domain from the input space. 
The inducing variables $\bu$, which we used to represent outputs of the GP at the inducing points, are now defined as $\bu = g(\bZ) := [g(\bz_1), \dots, g(\bz_M)]^\top$, where $g$ is a different function from $f$ that takes inputs in the domain of inducing points.
In convolutional GPs~\citep{van2017convolutional}, the input domain is the space of images, while the inducing points are in the space of image patches.
The convolutional GP function is defined as
$f(\bx) = \sum_{p} w_p g\left(\bx^{[p]}\right),$
where $g\sim \mathcal{GP}(0,k_g)$, $\bx^{[p]}$ is the $p$th patch in $\bx$, and $\bw = [w_1, \dots, w_P]^\top$ are the assigned weights for different patches.
In SOLVE-GP, we can choose either $\bZ$, $\bO$, or both to be inter-domain as long as we can compute the covariance between $\bu,\bv$ and $\bbf$.
For convolutional GPs, we let $\bZ$ and $\bO$ both be collections of image patches. Examples of the covariance matrices we need for this model include $\Kvf$ and $\Kvu$~(used for $\bC_{\bv\bv}$). They can be computed as
\begin{align*}
    [\Kvf]_{ij} &= \cov[g(\bo_i), f(\bx_j)] = \sum_{p} w_p k_g(\bo_i,\bx_j^{[p]}), \\
    [\Kvu]_{ij} &= \cov[g(\bo_i), g(\bz_j)] = k_g(\bo_i,\bz_j).
\end{align*}

\paragraph{Deep GPs.}
We show that we can integrate SOLVE-GP with popular doubly stochastic variational inference algorithms for deep GPs~\citep{salimbeni2017doubly}. The joint distribution of a deep GP model with inducing variables in all layers is
\begin{equation*}
    p(\by, \bbf^{1:L}, \bu^{1:L}) =  p(\by|\bbf^L)\prod_{\ell=1}^L \left[p(\bbf^\ell|\bu^\ell, \bbf^{\ell - 1})p(\bu^\ell)\right],
\end{equation*}
where we define $\bbf^0 = \bX$ and $\bbf^\ell$ is the output of the $\ell$th-layer GP. The doubly stochastic algorithm applies SVGP methods to each layer conditioned on samples from the variational distribution in the previous layer. The variational distribution over $\bu^{1:L},\bbf^{1:L}$ is
$
    q(\bbf^{1:L},\bu^{1:L}) = \prod_{\ell=1}^L\left[p(\bbf^\ell|\bu^\ell, \bbf^{\ell-1})q(\bu^\ell)\right].
$
This gives a similar objective as in the single layer case (Eq.~\eqref{eq:svgp}):
$
    \mathbb{E}_{q(\bbf^L)} \left[\log p(\by|\bbf^L)\right] - \sum_{\ell=1}^L\KL{q(\bu^\ell)}{p(\bu^\ell)},
$
where $q(\bbf^L) = \int \prod_{\ell=1}^L\left[p(\bbf^\ell|\bu^\ell, \bbf^{\ell-1})q(\bu^\ell)d\bu^\ell \right] d\bbf^{1:L-1}$. Extending this using SOLVE-GP is straightforward: we simply introduce orthogonal inducing variables $\vperp^{1:L}$ for all layers, which yields the lower bound:
\begin{align}
    &\mathbb{E}_{q(\bu^L, \fperp^L)}[\log p(\by|\fperp^L + \Kfu^L(\Kuu^L)^{-1}\bu^L)] - \notag \\ &\;\sum_{\ell=1}^L\left\{\mathrm{KL}[q(\bu^\ell)\|p(\bu^\ell)] +  \mathrm{KL}[q(\vperp^\ell)\|p_\perp(\vperp^\ell)]\right\}.
    \label{eq:solve-dgp}
\end{align}
The expression for $q(\bu^L,\fperp^L)$ is given in \cref{app:deep-gp}.

\section{RELATED WORK}

Many approximate algorithms have been proposed to overcome the computational limitations of GPs. 
The simplest of these are based on subsampling, such as the subset-of-data training~\citep{rasmussen2006gaussian} and the Nystr{\"o}m approximation~\citep{williams2001using}.
Better approximations can be constructed by learning a set of inducing points to summarize the dataset. 
As mentioned in \cref{sec:bg}, these works can be divided into approximations to the GP prior~\citep[SoR, DTC, FITC, etc.;][]{quinonero2005unifying}, and sparse variational methods~\citep{titsias2009variational,hensman2013gaussian,hensman2015scalable}.

Recently there have been many attempts to reduce the %
computational cost of 
using a large set of inducing points. 
A notable line of work~\citep{wilson2015kernel,evans2018scalable,gardner2018product} involves imposing grid structures on the locations of $\bZ$ to perform fast structure-exploiting computations. 
However, to get such benefits $\bZ$ need to be fixed due to the structure constraints, which often suffers from curse of dimensionality in the input space. 

Another direction for allowing the use of more inducing points is the decoupled method~\citep{cheng2017variational}, where two different sets of inducing points are used for modeling the mean and the covariance function. 
This gives linear complexity in the number of mean inducing points which allows using many more of them. 
Despite the increasing interest in decoupled inducing points~\citep{havasi2018deep,salimbeni2018orthogonally}, the method has not been well understood due to its complexity. 
We found that SOLVE-GP is closely related to a recent development of decoupled methods: the orthogonally decoupled variational GP~\citep[ODVGP,][]{salimbeni2018orthogonally}, as explained next.

\paragraph{Connection with Decoupled Inducing Points.}
If we set the $\beta$ and $\gamma$ inducing points in ODVGP~\citep{salimbeni2018orthogonally} to be $\bZ$ and $\bO$, 
their approach becomes equivalent to using the variational distribution $q'(\bu,\bv) = \mathcal{N}(\bbm'_{\bu,\bv},\bS'_{\bu,\bv})$, where
\begin{align*}
&\bbm'_{\bu,\bv} = \begin{bmatrix}
\bbm_\bu \\ \bbm_\bv + \bK_{\bv\bu}\bK_{\bu\bu}^{-1}\bbm_\bu
\end{bmatrix},\quad
\bS'_{\bu,\bv} = \\ 
&\begin{bmatrix} 
\bS_\bu \!&\! \bS_\bu\bK_{\bu\bu}^{-1}\bK_{\bu\bv} \\
\bK_{\bv\bu}\bK_{\bu\bu}^{-1}\bS_\bu \!&\! \bK_{\bv\bv} \!+\! \bK_{\bv\bu}\bK_{\bu\bu}^{-1}(\bS_\bu \!-\!\bK_{\bu\bu})\bK_{\bu\bu}^{-1}\bK_{\bu\bv}
\end{bmatrix}
.
\end{align*}
By comparing $\bS_{\bu,\bv}$ to $\bS'_{\bu,\bv}$, we can see that we generalize their method by introducing $\bS_\bv$, which replaces the original residual $\bK_{\bv\bv} - \bK_{\bv\bu}\bK_{\bu\bu}^{-1}\bK_{\bu\bv}$ (or $\Cvv$), so that we allow more flexible covariance modeling while still keeping the block structure. %
Thus ODVGP is a special case of SOLVE-GP where $q(\vperp)$ is restricted to have the same covariance $\Cvv$ as the prior.

\section{EXPERIMENTS}

Since ODVGP is a special case of SOLVE-GP, we use $M, M_2$ to refer to $|\beta|$ and $|\gamma|$ in their algorithm, respectively.

\subsection{1D Regression}

\begin{figure*}[t]%
    \begin{subfigure}[b]{0.244\linewidth}
        \centering
        \includegraphics[width=\textwidth]{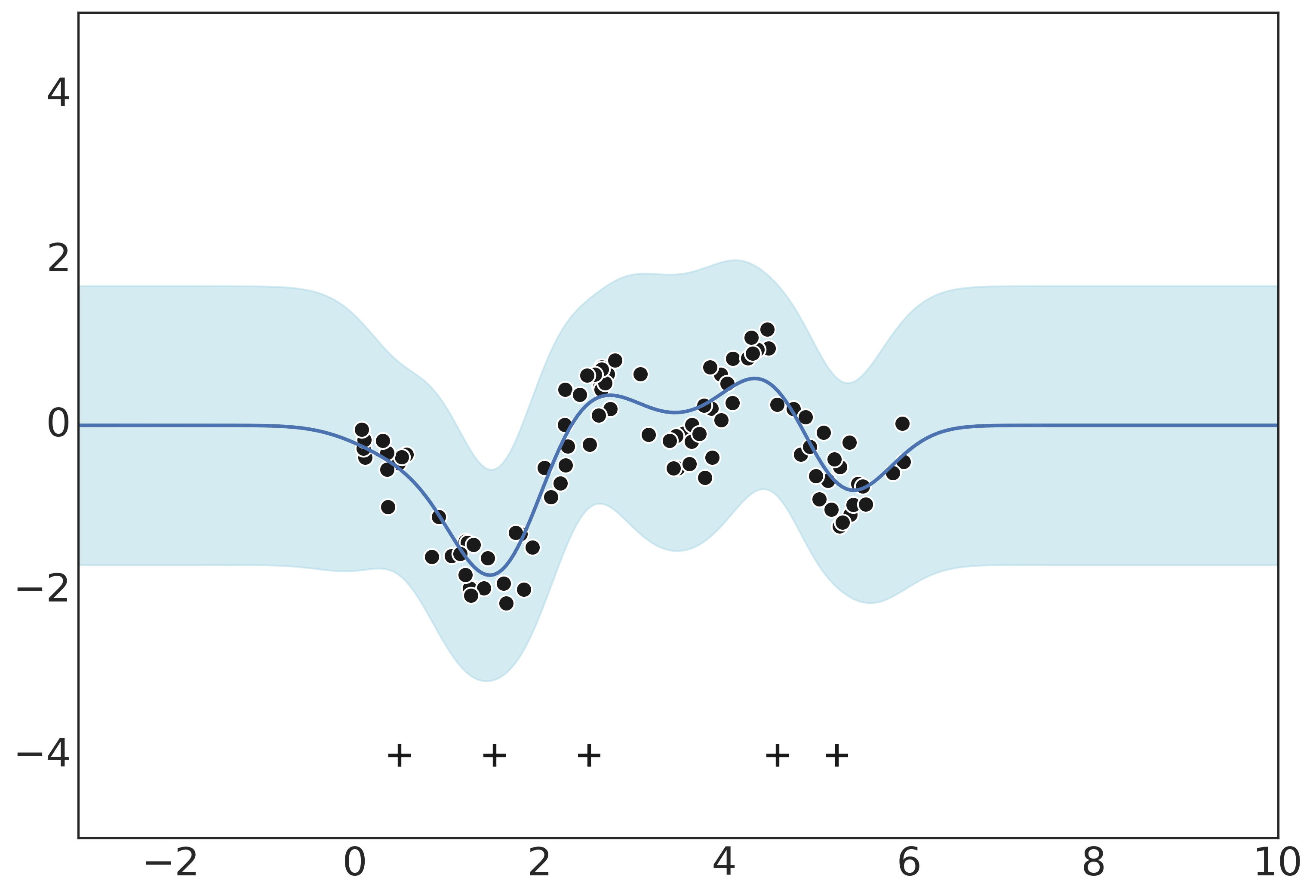}
        \caption{SVGP, $5$}
        \label{fig:toy-svgp-5}
    \end{subfigure}
    \begin{subfigure}[b]{0.244\linewidth}
        \centering
        \includegraphics[width=\textwidth]{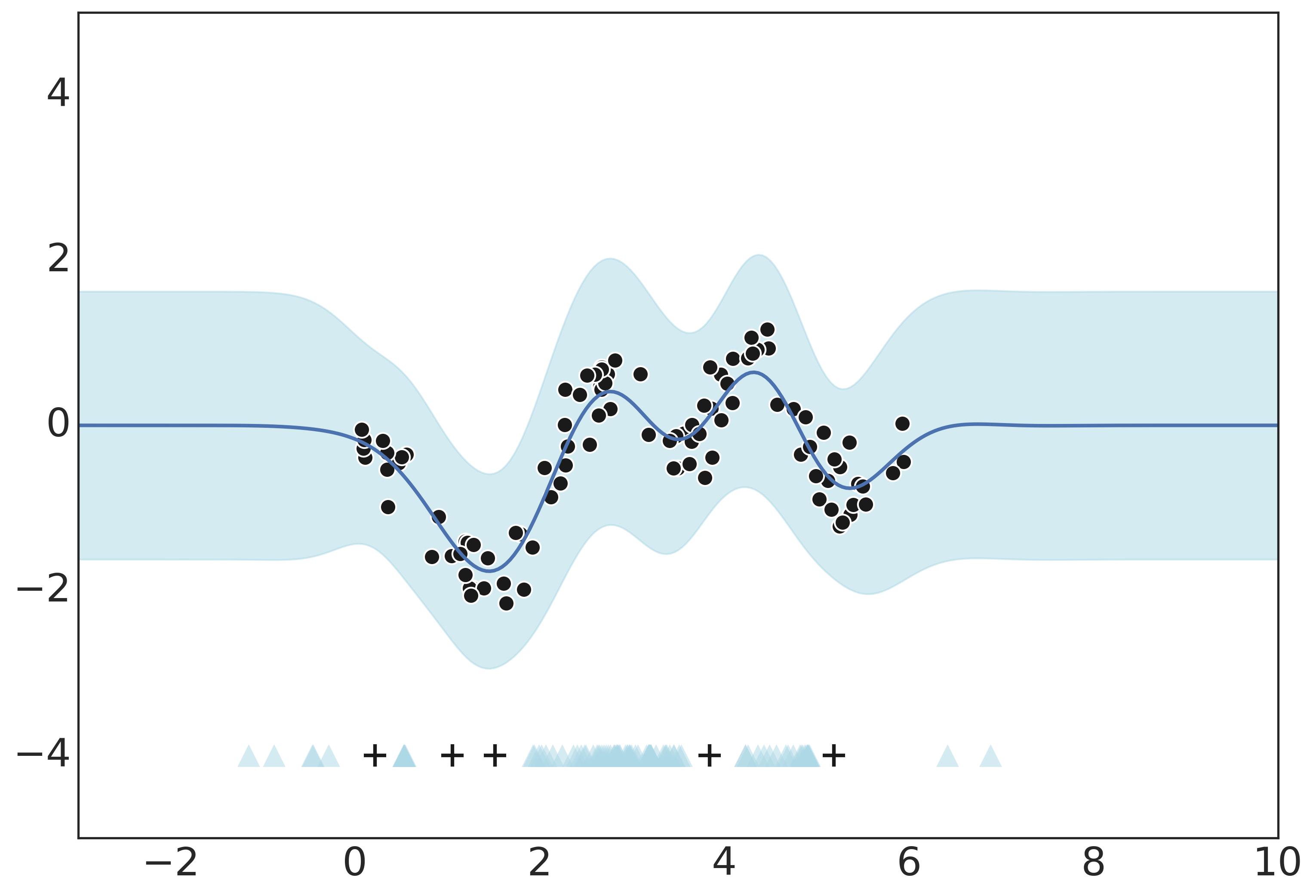}
        \caption{ODVGP, $5+100$}
        \label{fig:toy-odvgp}
    \end{subfigure}
    \begin{subfigure}[b]{0.244\linewidth}
        \centering
        \includegraphics[width=\textwidth]{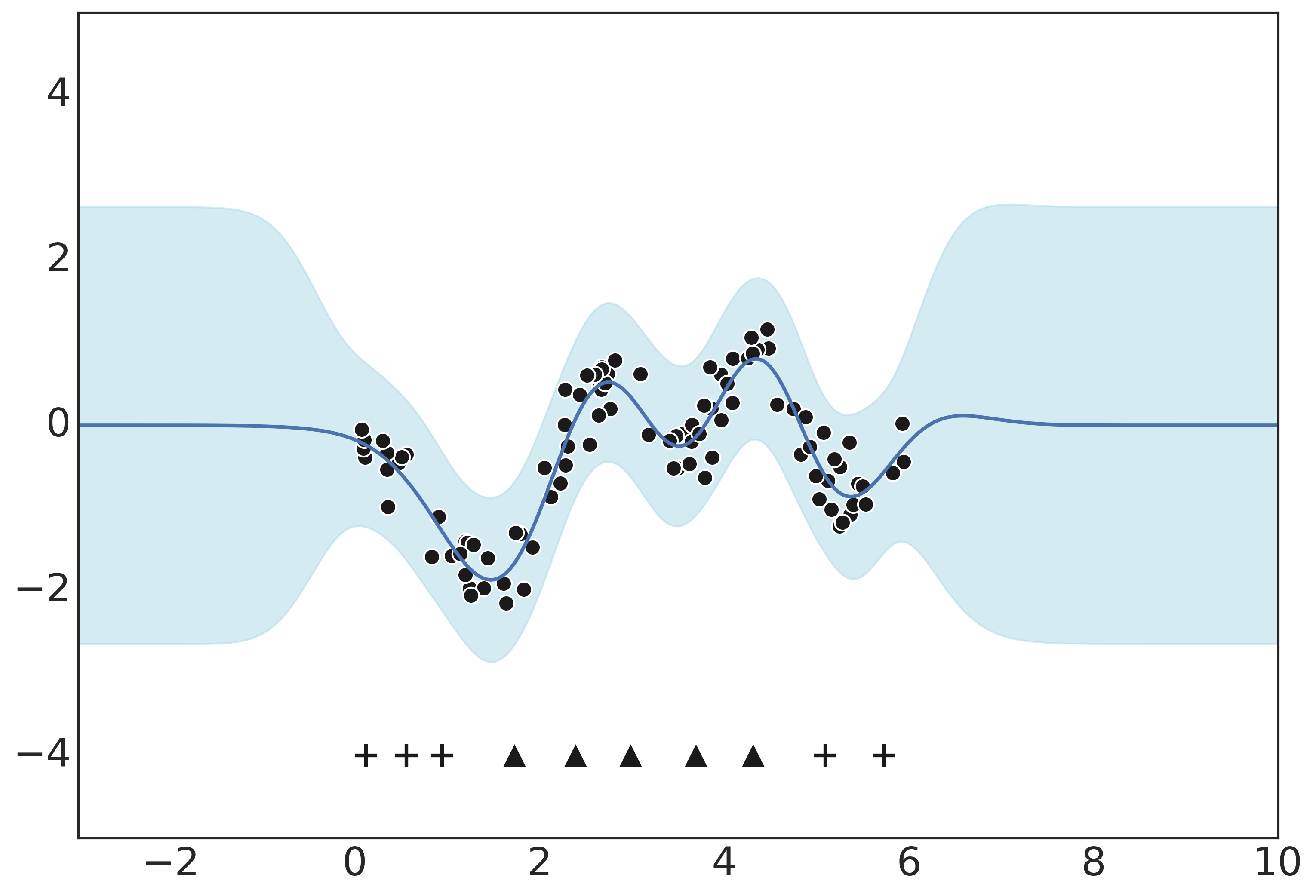}
        \caption{SOLVE-GP, $5+5$}
        \label{fig:toy-solvegp}
    \end{subfigure}
    \begin{subfigure}[b]{0.244\linewidth}
        \centering
        \includegraphics[width=\textwidth]{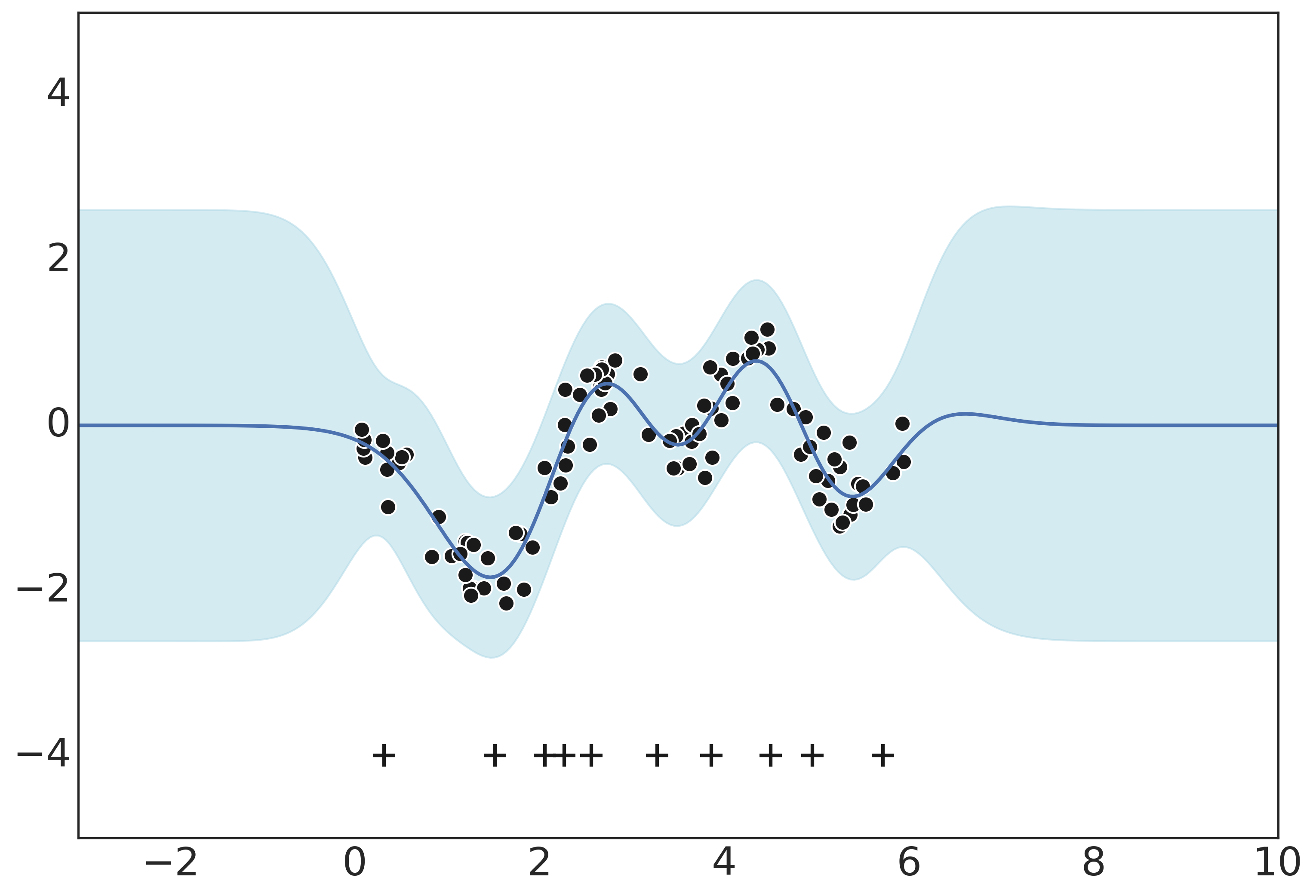}
        \caption{SVGP, $10$}
        \label{fig:toy-svgp-10}
    \end{subfigure}
    \caption{Posterior processes on the Snelson dataset, where shaded bands correspond to intervals of $\pm3$ standard deviations. The learned inducing locations are shown at the bottom of each figure, where $+$ correspond to $\bZ$; blue and dark triangles correspond to $\bO$ in ODVGP and SOLVE-GP, respectively.}
    \label{fig:snelson}
\end{figure*}

\begin{figure*}[t]%
    \begin{subfigure}[b]{0.49\textwidth}
        \centering
        \includegraphics[width=0.493\textwidth]{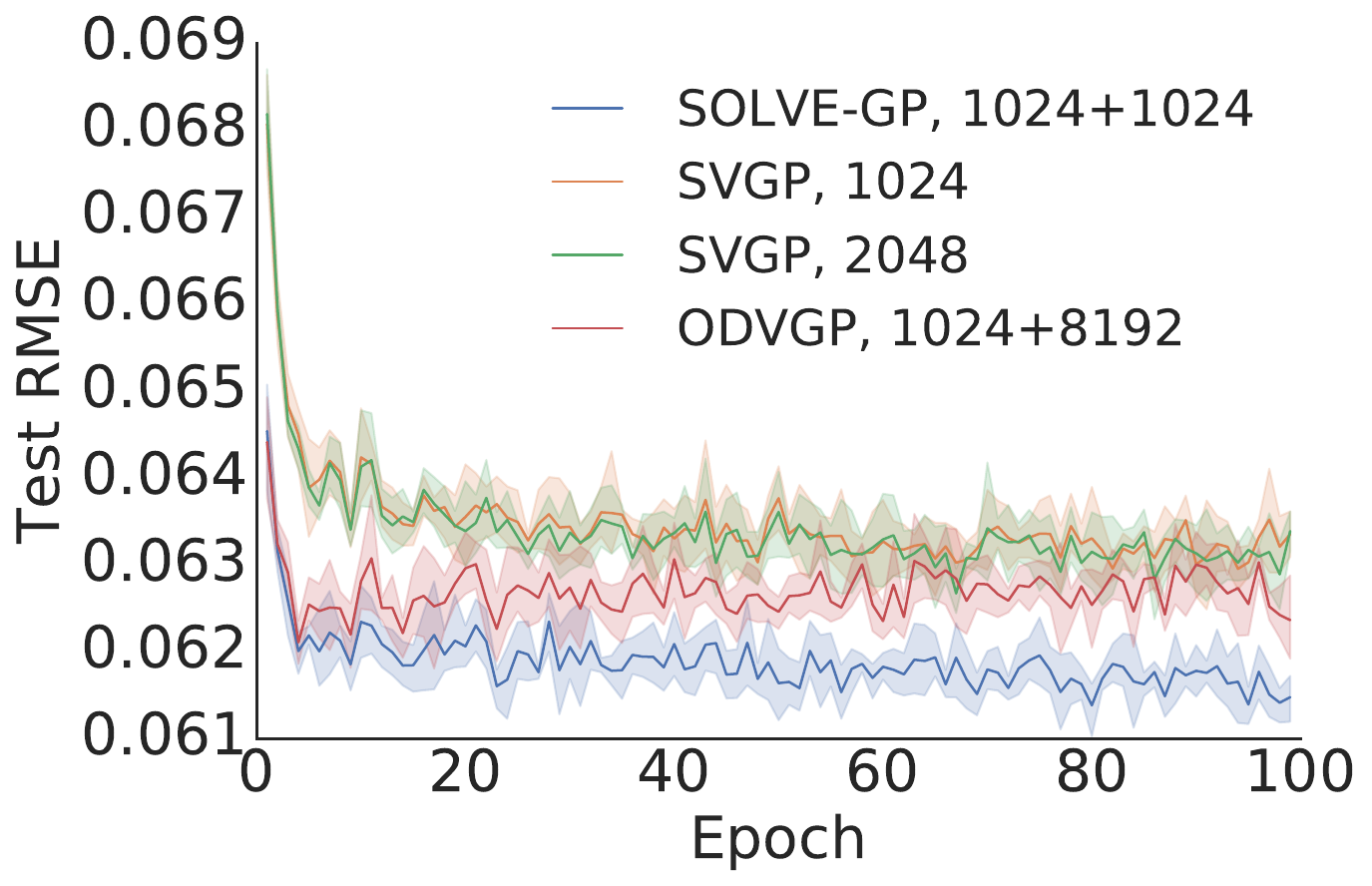}
        \includegraphics[width=0.493\textwidth]{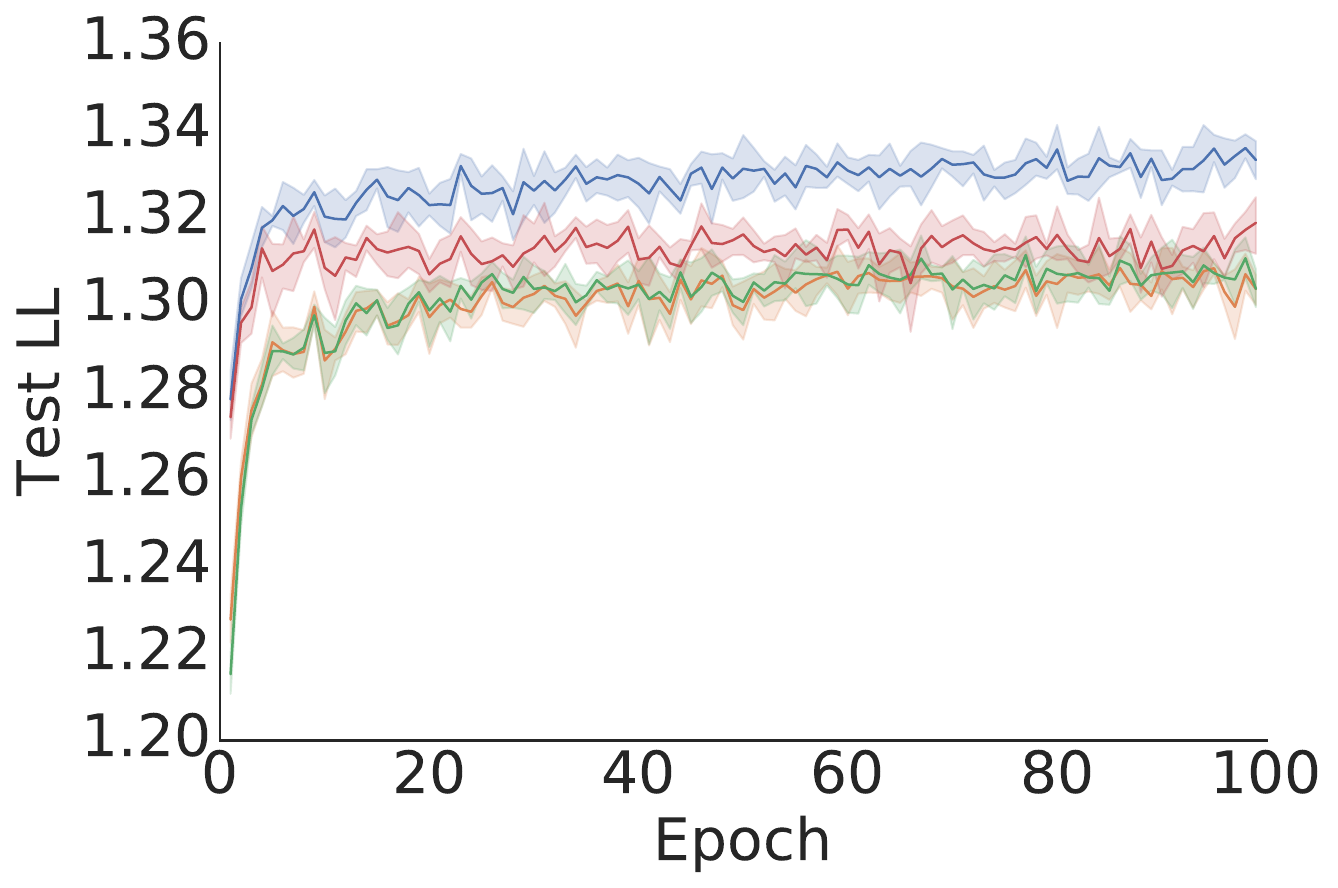}
        \caption{w/o whitening}
        \label{fig:house}
    \end{subfigure}
    \hspace{0.1em}
    \begin{subfigure}[b]{0.5\textwidth}
        \centering
        \includegraphics[width=0.493\textwidth]{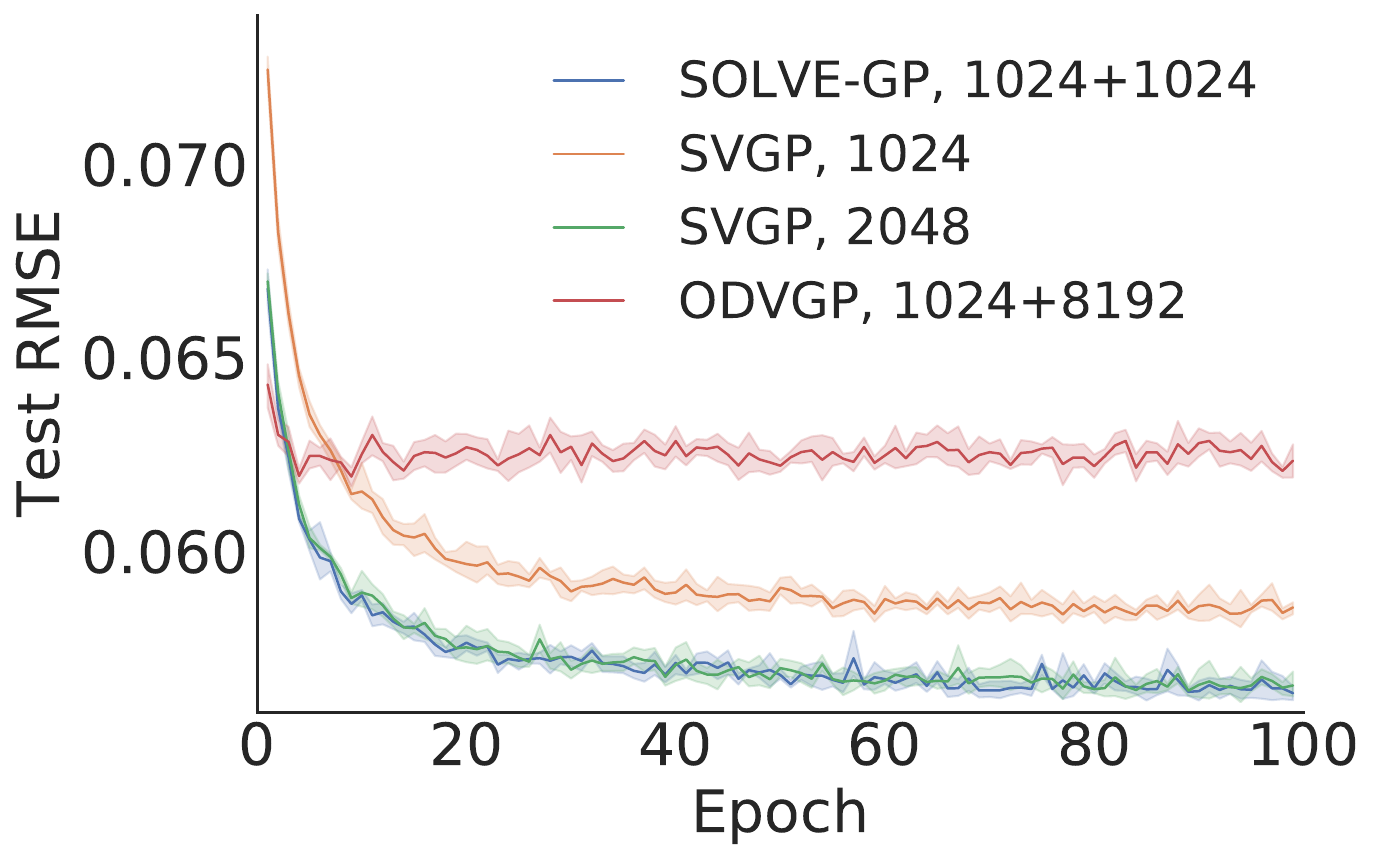}
        \includegraphics[width=0.493\textwidth]{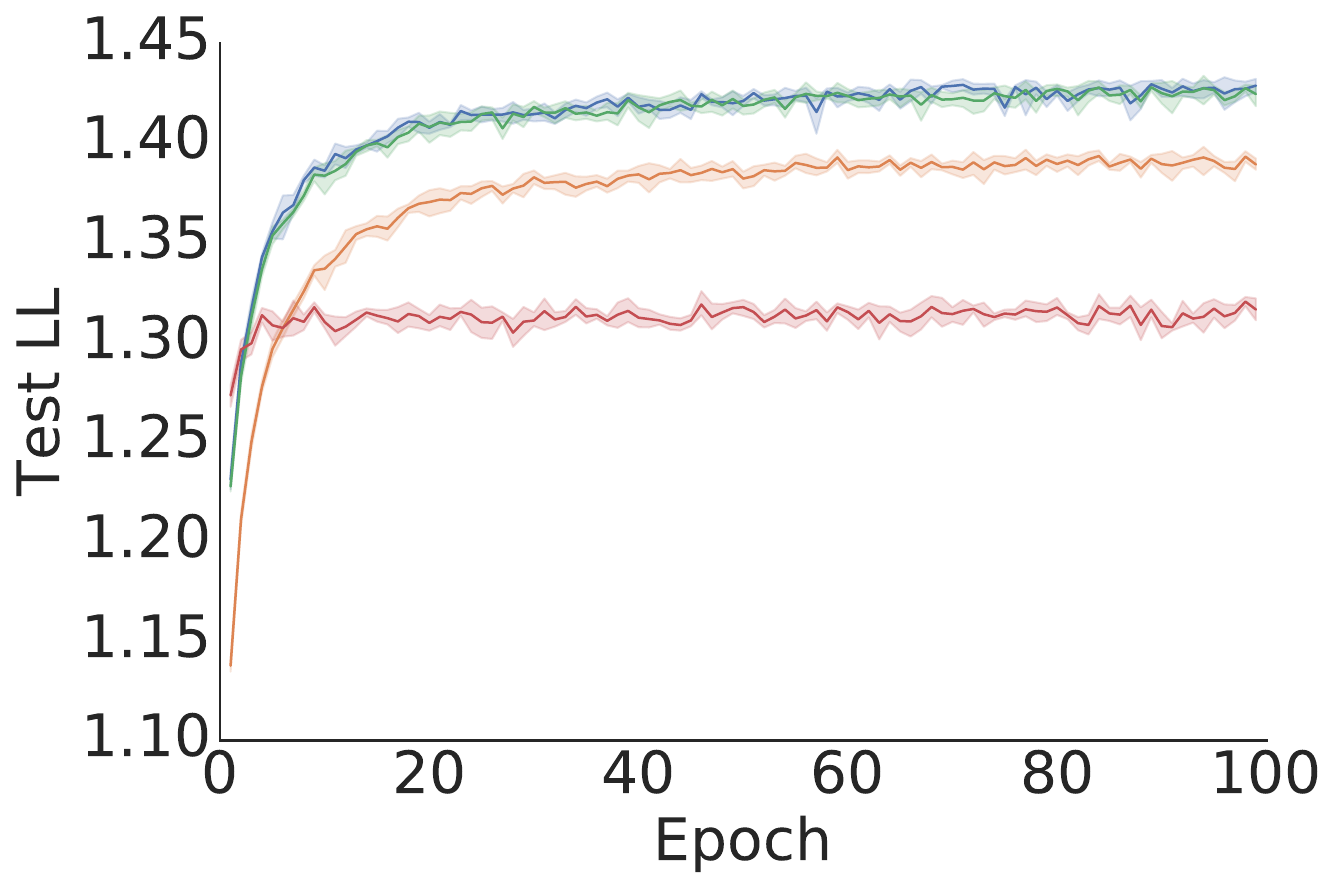}
        \caption{w/ whitening (except ODVGP)}
        \label{fig:house-white}
    \end{subfigure}
    \caption{Test RMSE and predictive log-likelihoods during training on HouseElectric.}
    \label{fig:regression}
\end{figure*}

We begin by illustrating our method on Snelson's 1D regression problem~\citep{snelson2006sparse} with 100 training points and mini-batch size 20. 
We compare the following methods: SVGP with 5 and 10 inducing points, ODVGP~($M=5, M_2=100$), and SOLVE-GP~($M=5, M_2=5$). 

The results are plotted in Fig.~\ref{fig:snelson}.
First we can see that 5 inducing points are insufficient to summarize the training set: SVGP~($M=5$) cannot fit data well and underestimates the variance in regions beyond the training data. 
Increasing $M$ to 10 fixes the issues, 
but requires 8x more computation for the Cholesky decomposition than using 5 inducing points\footnote{In practice the cost is negligible in this toy problem but we are analyzing the theoretical complexity.}. 
The decoupled formulation provides a cheaper alternative and we have tried ODVGP~($M=5, M_2=100$), which has 100 additional inducing points for modeling the mean function.
Comparing Fig.~\ref{fig:toy-svgp-5} and Fig.~\ref{fig:toy-odvgp}, we can see that this results in a much better fit for the mean function. 
However, the model still overestimates the predictive variance. 
As ODVGP is a special case of the SOLVE-GP framework, we can improve on it in terms of covariance modeling. 
As seen in Fig.~\ref{fig:toy-solvegp}, adding 5 orthogonal inducing points can closely approximate the results of SVGP~($M=10$), 
with only a 2-fold increase in the cost of the Cholesky decomposition relative to SVGP~($M=5$).

\subsection{Convolutional GP Models}
One class of applications that benefit from the SOLVE-GP framework is the training of large, hierarchical GP models where the true posterior distribution is difficult to approximate with a small number of inducing points. 
Convolutional GPs~\citep{van2017convolutional} and their deep variants~\citep{blomqvist2018deep,dutordoir2019translation} are such models. 
There inducing points are feature detectors just like CNN filters, which play a critical role in predictive performance.
As explained in \cref{sec:ext}, it is straightforward to apply SOLVE-GP to these models.

\paragraph{Convolutional GPs.} 
We train convolutional GPs on the CIFAR-10 dataset, using GPs with TICK kernels~\citep{dutordoir2019translation} to define the patch response functions. 
\Cref{tab:conv} shows the results for SVGP with 1K and 2K inducing points, SOLVE-GP~($M=1\text{K}, M_2=1\text{K}$), and  SVGP~($M=1.6\text{K}$) that has a similar running time on GPU as  SOLVE-GP. 
Clearly SOLVE-GP outperforms SVGP~($M=1\text{K}$). 
It also outperforms SVGP~($M=1.6\text{K}$), which has the same running time, and 
performs on par with the more expensive SVGP~($M=2\text{K}$), which is very encouraging.
This suggests that the structured covariance approximation is fairly accurate even for this large, non-conjugate model.

\paragraph{Deep Convolutional GPs.}
We further extend SOLVE-GP to deep convolutional GPs using the techniques described in \cref{sec:ext}. 
We experiment with 2-layer and 3-layer models that have 1K inducing points in the output layer and 384 inducing points in other layers.
The results are summarized in \Cref{tab:deep-conv}. 
These models are already quite slow to train on a single GPU, as indicated by the time per iteration.
SOLVE-GP allows to double the number of inducing points in each layer with only a 2-fold increase in computation.
This gives superior performance on both accuracy and test predictive likelihoods.
The double-size SVGP takes a week to run and is only for comparison purpose.

As shown above, on both single layer and deep convolutional GPs, we improve the state-of-the-art results of CIFAR-10 classification by 3-4 percentage points. 
This leads to more than $80\%$ accuracy on CIFAR-10 with a purely GP-based model, without any neural network components, closing the gap between GP/kernel regression and CNN baselines presented in \citet{novak2019bayesian,arora2019exact}. 
Note that all the results are obtained without data augmentation.

\begin{table}[t]
    \caption{Convolutional GPs for CIFAR-10 classification. Previous SOTA is 64.6\% by SVGP with 1K inducing points~\citep{van2017convolutional}.}
    \label{tab:conv}
    \resizebox{1.0\columnwidth}{!}{
    \centering
    \begin{tabular}{ l r cc c}
        \toprule
        & $M(+M_2)$ & Test Acc & Test LL & Time \\
        \midrule
        \multirow{2}{*}{SVGP}
        & $1\text{K}$  & 66.07\% & -1.59 & 0.241 s/iter \\
        & $1.6\text{K}$  & 67.18\% & -1.54 & 0.380 s/iter \\
        \multirow{1}{*}{SOLVE-GP} & $1\text{K}+1\text{K}$ & \bf 68.19\% & \bf -1.51 & 0.370 s/iter \\
        \midrule
        \multirow{1}{*}{\shortstack{SVGP}}
        & $2\text{K}$  & 68.06\% & \bf -1.48 & 0.474 s/iter \\
        \bottomrule
    \end{tabular}
    }
\end{table}

\begin{table*}[t]
\caption{Test log-likelihood values for the regression datasets. The numbers in parentheses are standard errors. Best mean values are highlighted, and asterisks indicate statistical significance.} 
\label{tab:ll}
\makebox[\textwidth][c]{
\resizebox{1.02\textwidth}{!}{
\setlength\tabcolsep{2pt}
\begin{tabular}{ l l cc cc cc cc}
\toprule
& & Kin40k &  Protein & \footnotesize KeggDirected & KEGGU &  3dRoad & Song &  Buzz & \footnotesize HouseElectric \\
\cmidrule(lr){3-10}
& $N$ & 25,600 & 29,267 & 31,248 & 40,708 & 278,319 & 329,820 & 373,280 & 1,311,539  \\
& $d$ & 8 & 9 & 20 & 27 & 3 & 90 & 77 & 9  \\
\midrule
\multirow{2}{*}{SVGP}
& $1024$  
& 0.094(0.003) & -0.963(0.006) & 0.967(0.005) & 0.678(0.004) & -0.698(0.002) & -1.193(0.001) & -0.079(0.002) & 1.304(0.002)  \\
& $1536$  
& 0.129(0.003) & -0.949(0.005) & 0.944(0.006) & 0.673(0.004) & -0.674(0.003) & -1.193(0.001) & -0.079(0.002) & 1.304(0.003) \\
\midrule
\multirow{2}{*}{ODVGP} & $1024+1024$ 
& 0.137(0.003) & -0.956(0.005) & -0.199(0.067) & 0.105(0.033) & -0.664(0.003) & -1.193(0.001) & -0.078(0.001) & 1.317(0.002) \\
& $1024+8096$  
& 0.144(0.002) & -0.946(0.005) & -0.136(0.063) & 0.109(0.033) & {\bf -0.657}(0.003) & -1.193(0.001) & -0.079(0.001) & 1.319(0.004) \\
\midrule
\footnotesize{SOLVE-GP} & $1024 + 1024$ 
& *{\bf 0.187}(0.002) & -0.943(0.005) & {\bf 0.973}(0.003) &  {\bf 0.680}(0.003) & -0.659(0.002) & {\bf -1.192}(0.001) & *{\bf -0.071}(0.001) & *{\bf 1.333}(0.003) \\
\midrule
SVGP
& $2048$
& 0.137(0.003) & {\bf -0.940}(0.005) & 0.907(0.003) & 0.665(0.004) & -0.669(0.002) & {\bf -1.192}(0.001) & -0.079(0.002) & 1.304(0.003) \\
\bottomrule
\end{tabular}
}
}
\end{table*}

\begin{table}[t]\vskip 0.03in
\caption{Deep convolutional GPs for CIFAR-10 classification. Previous SOTA is 76.17\% by a 3-layer model with 384 inducing points in all layers~\citep{dutordoir2019translation}.}
\label{tab:deep-conv}
\centering
\begin{subtable}{1.0\columnwidth}
\centering
\caption{2-layer model}
\smallskip
\resizebox{1\columnwidth}{!}{
    \begin{tabular}{ l ccc}
        \toprule
        & SVGP & SOLVE-GP & SVGP \\
        $M(+M_2)$ & $384,1\text{K}$ & $384+384$, $1\text{K}+1\text{K}$ & $768,2\text{K}$ \\
        \midrule
        Test Acc &   76.35\% & \bf 77.80\% & 77.46\%  \\
        Test LL &   -1.04 &  \bf -0.98  & \bf -0.98 \\
        \midrule
        Time &   0.392 s/iter & 0.657 s/iter & 1.104 s/iter \\
        \bottomrule
    \end{tabular}
    }

\end{subtable}

\par\bigskip
\begin{subtable}{1\columnwidth}
\centering
\caption{3-layer model}
\smallskip
\resizebox{1.00\columnwidth}{!}{
    \begin{tabular}{ l ccc}
        \toprule
        & SVGP & SOLVE-GP & SVGP \\
        $M(+M_2)$ & $384,384,1\text{K}$  & \makecell{$384+384,384+384$,\\$1\text{K}+1\text{K}$} & $768,768,2\text{K}$ \\
        \midrule
        Test Acc &  78.76\% & \bf 80.30\% & \bf 80.33\% \\
        Test LL &   -0.88 &  \bf -0.79  & -0.82 \\
        \midrule
        Time &   0.418 s/iter & 0.752 s/iter & 1.246 s/iter \\
        \bottomrule
    \end{tabular}
    }

\end{subtable}

\end{table}

\subsection{Regression Benchmarks}

Besides classification experiments, we evaluate our method on 10 regression datasets, with size ranging from tens of thousands to millions. 
The settings are followed from \citet{wang2019exact} and described in detail in \cref{app:exp-details}. 
We implemented SVGP with $M=1024 \& 2048$ inducing points, ODVGP and SOLVE-GP~($M=1024, M_2=1024$), as well as SVGP with $M=1536$ inducing points, which has roughly the same training time per iteration on GPU as the SOLVE-GP objective. 
An attractive property of ODVGP is that by restricting the covariance of %
$q(\vperp)$ to be the same as the prior covariance $\bC_{\bv\bv}$, it can use far larger $M_2$, because the complexity is linear with $M_2$ by sub-sampling the columns of $\Kvv$ for each gradient update. 
Thus for a fair comparison, we also include ODVGP~($M_2=8096$), where in each iteration 1024 columns of $\Kvv$ are sampled to estimate the gradient. 
Other experimental details are given in \cref{app:exp-details}.

We report the predictive log-likelihoods on test data in \Cref{tab:ll}. 
For space reasons, we provide the results on two small datasets (Elevators, Bike) in \cref{app:exp}. 
We can see that performance of SOLVE-GP is competitive with SVGP~($M=2048$) that involves 4x more expensive Cholesky decomposition. 
Perhaps surprisingly, despite using a less flexible covariance in the variational distribution,  SOLVE-GP often outperforms SVGP~($M=2048$).
We believe this is due to the optimization difficulties introduced by the $2048\times 2048$ covariance matrix and will test hypothesis on the HouseElectric dataset below.
On most datasets, using a large number of additional inducing points for modeling the mean function did improve the performance, as shown by the comparison between ODVGP~($M_2=1024$) and ODVGP~($M_2=8096$).
However, more flexible covariance modeling seems to be more important, as
SOLVE-GP outperforms ODVGP~($M_2=8096$) on all datasets except for 3dRoad.

In Fig.~\ref{fig:house} we plot the evolution of test RMSE and test log-likelihoods during training on HouseElectric.
Interestingly, %
ODVGP~($M_2=8096$) performs on par with SOLVE-GP early in training before falling behind it substantially.
The beginning stage is likely where the additional inducing points give good predictions but are not in the best configuration for maximizing the training lower bounds.
This phenomenon is also observed on Protein, Elevators, and Kin40k. 
We believe such mismatch between the training lower bound and predictive performance is caused by fixing the covariance matrix of $q(\vperp)$ to the prior covariance.
SVGP~($M=2048$) does not improve over SVGP~($M=1024$) and is outperformed by SOLVE-GP. 
Suggested above, this might be due to the difficulty of optimising large covariance matrices.
To verify this, we tried the ``whitening'' trick~\citep{murray2010slice,hensman2015mcmc}, described in  \cref{app:whiten}, which is often used to make optimization easier by reducing the correlation in the posterior distributions. 
As shown in Fig.~\ref{fig:house-white}, the performance of SVGP~($M=2048$) and SOLVE-GP becomes similar with whitening. 
We did not use whitening in ODVGP because it has a slightly different parameterization to allow sub-sampling $\Kvv$.

\section{CONCLUSION}

We proposed SOLVE-GP, a new variational inference framework for GPs using inducing points, that unifies and generalizes previous sparse variational methods. 
This increases the number of inducing points we can use for a fixed computational budget, which allows to improve performance of large, hierarchical GP models at a manageable computational cost.
Future work includes experiments on challenging datasets like ImageNet and investigating other ways to improve the variational distribution, as mentioned in \cref{sec:repara}.

\subsubsection*{Acknowledgements}

We thank Alex Matthews and Yutian Chen for helpful suggestions on improving the paper.

\bibliography{references}
\bibliographystyle{plainnat}

\appendix
\onecolumn

\section{Tighter Sparse Variational Bounds for GP Regression}
\label{app:other-bounds}

As mentioned in \cref{sec:repara}, another way to improve the variational distribution $q(\bu)p_\perp(\fperp)$ in SVGP is to make  $\bu$ and $\fperp$ dependent.
The best possible approximation of this type is obtained by the setting $q(\bu)$ to 
the optimal exact posterior conditional $q^*(\bu) = p(\bu|\fperp,\by)$.
The corresponding collapsed bound for GP regression can be derived by analytically 
marginalising out $\bu$ from the joint model in Eq.~\eqref{eq:joint1}, 
\begin{align}
p(\by|\fperp) &= \int p(\by|\fperp + \bK_{\bbf\bu}\bK_{\bu\bu}^{-1}\bu)p(\bu)~d\bu \notag\\
&= \mathcal{N}(\by|\fperp,\bQ_{\bbf\bbf} + \sigma^2\bI),
\end{align}
and then forcing the approximation $p_\perp(\fperp)$:
\begin{equation}
\mathbb{E}_{p_\perp(\fperp)}\log \mathcal{N}(\by|\fperp,\bQ_{\bbf\bbf} + \sigma^2\bI).
\end{equation}
This bound has a closed-form as
\begin{equation}
    \log \mathcal{N}(\by|\bzero, \Qff + \sigma^2\bI) - \frac{1}{2}\mathrm{tr}\left[(\Qff + \sigma^2\bI)^{-1}(\Kff - \Qff)\right],
\end{equation}
Applying the matrix inversion lemma to $(\Qff + \sigma^2\bI)^{-1}$, we have an equivalent form that can be directly compared with Eq.~\eqref{eq:referencebound}:
\begin{equation}
    \underbrace{\log \mathcal{N}(\by|\bzero, \Qff + \sigma^2\bI) - \frac{1}{2\sigma^2}\mathrm{tr}(\Kff - \Qff)}_{=\text{Eq.~\eqref{eq:referencebound}}} + \frac{1}{2\sigma^{4}}\mathrm{tr}\left[\bK_{\bbf\bu}(\bK_{\bu\bu} + \sigma^{-2} \bK_{\bu\bbf} \bK_{\bbf\bu})^{-1} \bK_{\bu\bbf} (\Kff - \Qff)\right],
\end{equation}
where the first two terms recover Eq.~\eqref{eq:referencebound}, suggesting this is a tighter bound than the \citet{titsias2009variational} bound. 
This bound is not amenable to large-scale datasets because of $O(N^2)$ storage  and $O(M N^2)$ computation 
time (dominated by the matrix multiplication $\bK_{\bu\bbf} \Kff$) requirements. However, it is still of theoretical interest and can be applied to medium-sized regression datasets, just like the SGPR algorithm using the \citet{titsias2009variational} bound.

\section{Details of Orthogonal Decomposition}
\label{app:inner-product}

In \cref{sec:orth} we described the following orthogonal decomposition for $f\in\mathcal{H}$, where $\mathcal{H}$ is the RKHS induced by kernel $k$:
\begin{equation}
\label{eq:orth-begin}
f = f_\| + f_\perp,\quad f_\| \in V\text{ and }f_\perp \perp V.
\end{equation}
Here $V$ is the subspace spanned by the inducing basis: $V = \{\sum_{j=1}^M \alpha_j k(\bz_j, \cdot),\; \balpha = [\alpha_1,\dots, \alpha_M]^\top \in \mathbb{R}^M\}.$ 
Since $f_\| \in V$, we let $f_\| = \sum_{j=1}^M \alpha'_j k(\bz_j, \cdot)$. According to the properties of orthogonal projection, we have
\begin{equation}
	\label{eq:orth-proj}
	\langle f, g \rangle_\mathcal{H} = \langle f_\|, g \rangle_\mathcal{H}, \quad \forall g \in V,
\end{equation}
where $\langle\rangle_\mathcal{H}$ is the RKHS inner product that satisfies the reproducing property: $\langle f, k(\bx, \cdot)\rangle_\mathcal{H} = f(\bx)$.
Similarly let $g = \sum_{j=1}^M\beta_j k(\bz_j, \cdot)$. 
Then $\langle f, g \rangle_\mathcal{H} = \sum_{j=1}^M\beta_j f(\bz_j)$, $\langle f_\parallel, g \rangle_\mathcal{H} = \sum_{i=1}^M\sum_{j=1}^M \alpha'_i\beta_j k(\bz_i, \bz_j)$. 
Plugging into Eq.~\eqref{eq:orth-proj} and rearranging the terms, we have
\begin{equation}
	\bbeta^\top(f(\bZ) - k(\bZ,\bZ)\balpha') = 0,\quad \forall \bbeta \in \mathbb{R}^M,
\end{equation}
where $f(\bZ) = [f(\bz_1), \dots, f(\bz_M)]^\top$, $k(\bZ,\bZ)$ is a matrix with the $ij$-th term as $k(\bz_i, \bz_j)$, and $\balpha' = [\alpha'_1, \dots, \alpha'_M]^\top$. 
Therefore,
\begin{equation}
\label{eq:orth-end}
\balpha' = k(\bZ,\bZ)^{-1}f(\bZ),
\end{equation}
and it follows that $f_\|(\bx) = k(\bx, \bZ)k(\bZ,\bZ)^{-1}f(\bZ)$, where $k(\bx,\bZ) = [k(\bz_1, \bx), \dots, k(\bz_M,\bx)]$.
The above analysis arrives at the decomposition:
\begin{equation}
\label{eq:orth-decomp}
	f_\| = k(\cdot,\bZ)k(\bZ,\bZ)^{-1}f(\bZ),\quad f_\perp = f - f_\|.
\end{equation}
Although the derivation from Eq.~\eqref{eq:orth-begin} to \eqref{eq:orth-end} relies on the fact $f\in \mathcal{H}$, such that the inner product is well-defined, the decomposition in Eq.~\eqref{eq:orth-decomp} is valid for any function $f$ on $\mathcal{X}$.
This motivates us to study it for $f\sim \mathcal{GP}(0, k)$.
Substituting $\bu$ for $f(\bZ)$ and $\Kuu$ for $k(\bZ,\bZ)$, we have for $f\sim \mathcal{GP}(0,k)$:
\begin{align}
	f_\parallel = k(\cdot, \bZ)\Kuu^{-1}\bu \sim p_\parallel &\equiv \mathcal{GP}(0, k(\bx,\bZ)\Kuu^{-1}k(\bZ,\bx')), \\
	f_{\perp} \sim p_\perp &\equiv \mathcal{GP}(0, k(\bx,\bx') - k(\bx,\bZ)\Kuu^{-1}k(\bZ,\bx')).
\end{align}

\section{The Collapsed SOLVE-GP Lower Bound}
\label{app:collapsed-solvegp}

We derive the collapsed SOLVE-GP lower bound in Eq.~\eqref{eq:collapsed2} by seeking the optimal $q(\bu)$ that is independent of $\fperp$. 
First we rearrange the terms in the uncollapsed SOLVE-GP bound~(Eq.~\eqref{eq:sovgp}) as
\begin{equation} \label{eq:collapsed-step-1}
    \mathbb{E}_{q(\bu)}\left\{\mathbb{E}_{q_\perp(\fperp)}\left[\log \mathcal{N}\left(\by|\fperp + \Kfu\Kuu^{-1}\bu, \sigma^2\bI\right)\right] \right\}- \KL{q(\bu)}{p(\bu)} - \KL{q(\vperp)}{p_\perp(\vperp)}.
\end{equation}
where $q_\perp(\fperp) = \mathcal{N}(\bbm_{\fperp}, \bS_{\fperp})$, and $\bbm_{\fperp} = \Cfv\Cvv^{-1}\bbm_{\bv}$, $\bS_{\fperp}=\Cff + \Cfv\Cvv^{-1}(\bS_\bv - \Cvv)\Cvv^{-1}\Cvf$.
In the first term we can simplify the expectation over $\fperp$ as:
\begin{align}
    &\mathbb{E}_{q_\perp(\fperp)} \log \mathcal{N}\left(\by|\fperp + \Kfu\Kuu^{-1}\bu, \sigma^2\bI\right) \notag\\
    =\;& \mathbb{E}_{q_\perp(\fperp)} \left[-\frac{N}{2}\log 2\pi - \frac{N}{2}\log \sigma^2 - \frac{1}{2\sigma^2}(\by - \fperp - \Kfu\Kuu^{-1}\bu)^\top(\by - \fperp - \Kfu\Kuu^{-1}\bu)\right] \notag\\
    =\;& \left[-\frac{N}{2}\log 2\pi - \frac{N}{2}\log \sigma^2 - \frac{1}{2\sigma^2}(\by - \bbm_{\fperp} - \Kfu\Kuu^{-1}\bu)^\top(\by - \bbm_{\fperp} - \Kfu\Kuu^{-1}\bu)\right] \notag\\
    &\; - \mathbb{E}_{q_\perp(\fperp)}\left[\frac{1}{2\sigma^2}(\fperp - \bbm_{\fperp})^\top(\fperp - \bbm_{\fperp})\right] \notag\\
    =\;& \log \mathcal{N}(\by|\Kfu\Kuu^{-1}\bu + \bbm_{\fperp}, \sigma^2\bI) - \frac{1}{2\sigma^2}\mathrm{tr}(\bS_{\fperp}).
\end{align}
Plugging into Eq.~\eqref{eq:collapsed-step-1} and rearranging the terms, we have
\begin{equation}
    \underbrace{\mathbb{E}_{q(\bu)}\left[\log \mathcal{N}(\by|\Kfu\Kuu^{-1}\bu + \bbm_{\fperp}, \sigma^2\bI)\right] - \KL{q(\bu)}{p(\bu)}}_{\leq\log \int \mathcal{N}(\by|\Kfu\Kuu^{-1}\bu + \bbm_{\fperp}, \sigma^2\bI)p(\bu)\;d\bu} - \frac{1}{2\sigma^2}\mathrm{tr}(\bS_{\fperp}) - \KL{q(\vperp)}{p_\perp(\vperp)}.
\end{equation}
Clearly the leading two terms form a variational lower bound of the joint distribution $\mathcal{N}(\by|\Kfu\Kuu^{-1}\bu + \bbm_{\fperp}, \sigma^2\bI)p(\bu)$.
The optimal $q(\bu)$ will turn it into the log marginal likelihood:
\begin{equation}
    \log \int \mathcal{N}(\by|\Kfu\Kuu^{-1}\bu + \bbm_{\fperp}, \sigma^2\bI)p(\bu)\;d\bu = \log \mathcal{N}(\by|\bbm_{\fperp}, \Qff + \sigma^2\bI).
\end{equation}
Plugging this back, we have the collapsed SOLVE-GP bound in Eq.~\eqref{eq:collapsed2}:
\begin{equation}
\log \mathcal{N}(\by | \bC_{\bbf \bv} \bC_{\bv \bv}^{-1} \bbm_\bv,  \bQ_{\bbf\bbf} + \sigma^2 \bI)
-\frac{1}{2\sigma^2}\mathrm{tr}(\bS_{\fperp})
- \KL{\mathcal{N}(\bbm_\bv,\bS_\bv)}{\mathcal{N}(\bzero,\bC_{\bv\bv})},
\end{equation}
Moreover, we could find the optimal $q^*(\bv) = \mathcal{N}(\bbm_\bv^*, \bS_\bv^*)$ by setting the derivatives w.r.t. $\bbm_\bv$ and $\bS_\bv$ to be zeros:
\begin{align}
    \bbm_\bv^* &= \Cvv [\Cvv + \Cvf \bA^{-1} \Cfv]^{-1} \Cvf \bA^{-1} \by, \\
    \bS_\bv^* &= \Cvv [ \Cvv + \sigma^{-2} \Cvf \Cfv ]^{-1} \Cvv,
\end{align}
where $\bA = \Qff + \sigma^2 \bI$. Then the collapsed bound with the optimal $q(\vperp)$ is
\begin{equation}
    \log N(\by|\Cfv \Cvv^{-1} \bbm_\bv^*, \bA) - \frac{1}{2\sigma^2} \mathrm{tr}[\Cff - \bB(\bB+\sigma^2 \bI)^{-1}\bB] - \KL{\mathcal{N}(\bbm_\bv^*, \bS_\bv^*)}{\mathcal{N}(\bzero, \Cvv)},
\end{equation}
where $\bB = \Cfv \Cvv^{-1} \Cvf$.

\section{Computational Details}
\label{app:pred}

\subsection{Training} 
To compute the lower bound in Eq.~\eqref{eq:sovgp}, we write it as
\begin{equation}
\label{eq:comp1} 
	\sum_{n=1}^N\mathbb{E}_{q(f(\bx_n);\Theta)} [\log p(y_n|f(\bx_n))] - \KL{q(\bu)}{p(\bu)} {\color{blue}\;-\;\KL{q(\vperp)}{p_\perp(\vperp)}},
\end{equation}
where $\Theta := \{\bbm_\bu,\bS_\bu,\bbm_\bv,\bS_\bv,\bZ,\bO\}$ and $q(f(\bx_n);\Theta)$ defines the marginal distribution of $\bbf = \fperp + \Kfu\Kuu^{-1}\bu$ for the $n$-th data point given $\bu \sim q(\bu)$ and $\fperp \sim q_\perp(\fperp)$. 
We can write $q(f(\bx_n); \Theta)$ as
\begin{equation}
\label{eq:post-pred-single}
	q(f(\bx_n); \Theta) = \mathcal{N}(\mu(\bx_n), \sigma^2(\bx_n)),
\end{equation}
where
\begin{align}
\mu(\bx_n) &= k(\bx_n,\bZ)\Kuu^{-1}\bbm_\bu {\color{blue}\;+\;c(\bx_n, \bO)\Cvv^{-1}\bbm_{\bv}}, \label{eq:comp2}\\
\sigma^2(\bx_n) &=  k(\bx_n,\bZ)\Kuu^{-1}\bS_\bu\Kuu^{-1}k(\bZ,\bx_n) + c(\bx_n,\bx_n) {\color{blue}\;+\;c(\bx_n,\bO)\Cvv^{-1}(\bS_\bv - \Cvv)\Cvv^{-1}c(\bO,\bx_n)}. \label{eq:comp3}
\end{align}
Here $c(\bx,\bx') := k(\bx,\bx') - k(\bx,\bZ)\Kuu^{-1}k(\bZ,\bx)$ denotes the covariance function of $p_\perp$. 
The univariate expectation of $\log p(y_n|f(\bx_n))$ under $q(f(\bx_n);\Theta)$ can be computed in closed form (e.g., for Gaussian likelihoods) or using quadrature~\citep{hensman2015mcmc}.
It can also be estimated by Monte Carlo with the reparameterization trick~\citep{kingma2013auto,titsias2014doubly,rezende2014stochastic} to propagate gradients. 
For large datasets, an unbiased estimate of the sum can be used for mini-batch training: $\frac{N}{|B|}\sum_{(\bx,y)\in B}\mathbb{E}_{q(f(\bx);\Theta)}[\log p(y|f(\bx))]$, where $B$ denotes a small batch of data points.

Besides the log-likelihood term, we need to compute the two KL divergence terms:
\begin{align}
\KL{q(\bu)}{p(\bu)} &= 
\frac{1}{2}\left[\log\det\Kuu - \log\det\bS_\bu - M + \mathrm{tr}(\Kuu^{-1}\bS_\bu) + \bbm_\bu^\top\Kuu^{-1}\bbm_\bu \right], \\
{\color{blue}\KL{q(\vperp)}{p_\perp(\vperp)}} &{\color{blue}\;=
\frac{1}{2}\left[\log\det\Cvv - \log\det\bS_\bv - M + \mathrm{tr}(\Cvv^{-1}\bS_\bv) + \bbm_\bv^\top\Cvv^{-1}\bbm_\bv \right]}.
\end{align}

We note that if the blue parts in Eqs.~\eqref{eq:comp1}~to~\eqref{eq:comp3} are removed, then we recover the SVGP lower bound in Eq.~\eqref{eq:svgp}. 
An implementation of the above computations using the Cholesky decomposition is shown in \cref{alg:train}.

\begin{algorithm}[t]
	\caption{The SOLVE-GP lower bound via Cholesky decomposition. We parameterize the variational covariance matrices with their Cholesky factors $\bS_\bu=\Lu\Lu^\top,\bS_\bv=\Lv\Lv^\top$. $\bA=\Lu^0\setminus\Kuv$ denotes the solution of $\Lu^0\bA=\Kuv$.  $\odot$ denotes elementwise multiplication. The differences from SVGP are shown in {\color{blue}blue}.}
	\label{alg:train}
	\begin{algorithmic}[1]
		\Require $\bX$ (training inputs), $\by$ (targets), $\bZ,\bO$ (inducing points), $\bbm_\bu,\Lu,\bbm_\bv,\Lv$ (variational parameters)
		\State $\Kuu = k(\bZ,\bZ)$, {\color{blue}$\Kvv = k(\bO,\bO)$}
		\State $\Lu^0 = \uline{\mathrm{Cholesky}(\Kuu)}$, {\color{blue}$\Kuv = k(\bZ,\bO)$, $\bA := \uline{\Lu^0\setminus\Kuv}$, $\Cvv = \Kvv - \uline{\bA^\top\bA}$, $\Lv^0 = \uline{\mathrm{Cholesky}(\Cvv)}$}
		\State $\Kuf = k(\bZ,\bX)$, {\color{blue}$\Kvf = k(\bO,\bX)$}
		\State $\bB := \uline{\Lu^0 \setminus \Kuf}$, {\color{blue}$\Cvf = \Kvf - \uline{\bA^\top \bB}$, $\bD := \uline{\Lv^0 \setminus \Cvf}$}
		\State $\bE := \uline{(\Lu^0)^\top \setminus \bB}$, $\bF := \uline{\Lu^\top \bE}$, {\color{blue}$\bG := \uline{(\Lv^0)^\top \setminus \bD}$, $\bH := \uline{\Lv^\top \bG}$}
		\State $\bmu(\bX) = \bE^\top \bbm_\bu {\color{blue}\;+\;\bG^\top \bbm_\bv}$
		\State $\bsigma^2(\bX) = \diag(\Kff) +  (\bF\odot\bF)^\top\mathbf{1} - (\bB \odot \bB)^\top\mathbf{1} {\color{blue}\;+\;(\bH\odot\bH)^\top\mathbf{1} - (\bD\odot\bD)^\top\mathbf{1}}$
		\State Compute $\mathrm{LLD} = \sum_{n=1}^N\mathbb{E}_{\mathcal{N}(\bmu(\bx_n), \bsigma^2(\bx_n))} \log p(y_n|f(\bx_n))$ in closed form or using quadrature/Monte Carlo.
		\Function{Compute\_KL}{$\bbm$, $\bL$, $\bL^0$}
		\State $\bP = \uline{\bL^0\setminus\bL}$, $\ba = \bL^0 \setminus \bbm$
		\State \Return $\log(\diag(\bL^0))^\top\mathbf{1} - \log(\diag(\bL))^\top\mathbf{1} + 1/2((\bP\odot\bP)^\top\mathbf{1} + \ba^\top\ba - M)$
		\EndFunction
		\State $\mathrm{KL}_{\bu} = \Call{Compute\_KL}{\bbm_\bu, \Lu, \Lu^0}$, {\color{blue}$\mathrm{KL}_{\bv} = \Call{Compute\_KL}{\bbm_\bv, \Lv, \Lv^0}$}
		\State \Return $\mathrm{LLD} - \mathrm{KL}_{\bu} {\color{blue}- \mathrm{KL}_{\bv}}$
	\end{algorithmic}
\end{algorithm}

\subsection{Prediction}
We can predict the function value at a test point $\bx^*$ with the approximate posterior by substituting $\bx^*$ for $\bx_n$ in Eq.~\eqref{eq:post-pred-single}.
For multiple test points $\bX^*$, we denote the joint predictive density by $\mathcal{N}(\bbf^*|\bmu^{*}, \bSigma^{*}) %
$,
where %
the predicted mean and covariance are
\begin{align}
\bmu^* &= \bK_{*\bu}\bK_{\bu\bu}^{-1}\bbm_\bu {\color{blue} + \bC_{*\bv}\bC_{\bv\bv}^{-1}\bbm_\bv}, \\
\bSigma^* &= \bK_{*\bu}\bK_{\bu\bu}^{-1}\bS_\bu\bK_{\bu\bu}^{-1}\bK_{\bu*} + \bC_{**} {\color{blue} - \bC_{*\bv}\bC_{\bv\bv}^{-1}(\bC_{\bv\bv} - \bS_\bv)\bC_{\bv\bv}^{-1}\bC_{\bv*}}.
\end{align}

\subsection{Computational Complexity}

As mentioned in \cref{sec:solvegp-lb}, the time complexity of SOLVE-GP is $\mathcal{O}(N\bar{M}^2 + \bar{M}^3)$ per gradient update, where $\bar{M}=\max(M,M_2)$ and $N$ is the batch size.
Here we provide a more fine-grained analysis by counting cubic-cost operations and compare to the standard SVGP method. 
We underlined all the cubic-cost operations in \cref{alg:train}, including matrix multiplication, Cholesky decomposition, and solving triangular matrix equations.
We count them for SVGP and SOLVE-GP. The results are summarized in \Cref{tab:cost}.

\begin{figure}[tbp] %
	\centering
	\begin{subfigure}[b]{0.4\linewidth}
		\centering
		\includegraphics[height=5cm]{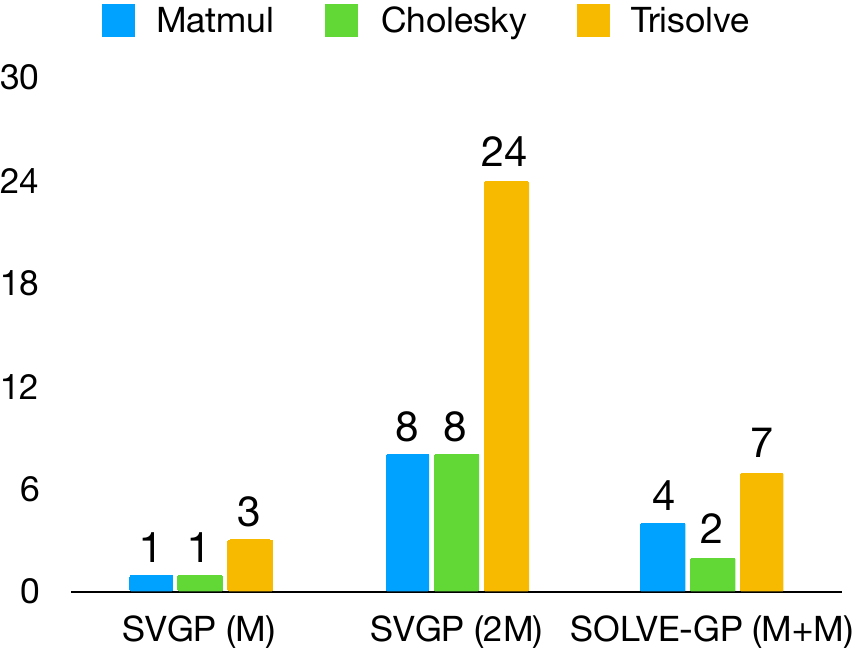}
		\caption{$N\approx M$}
		\label{fig:cost1}
	\end{subfigure}
\quad
	\begin{subfigure}[b]{0.4\linewidth}
		\centering
		\includegraphics[height=5cm]{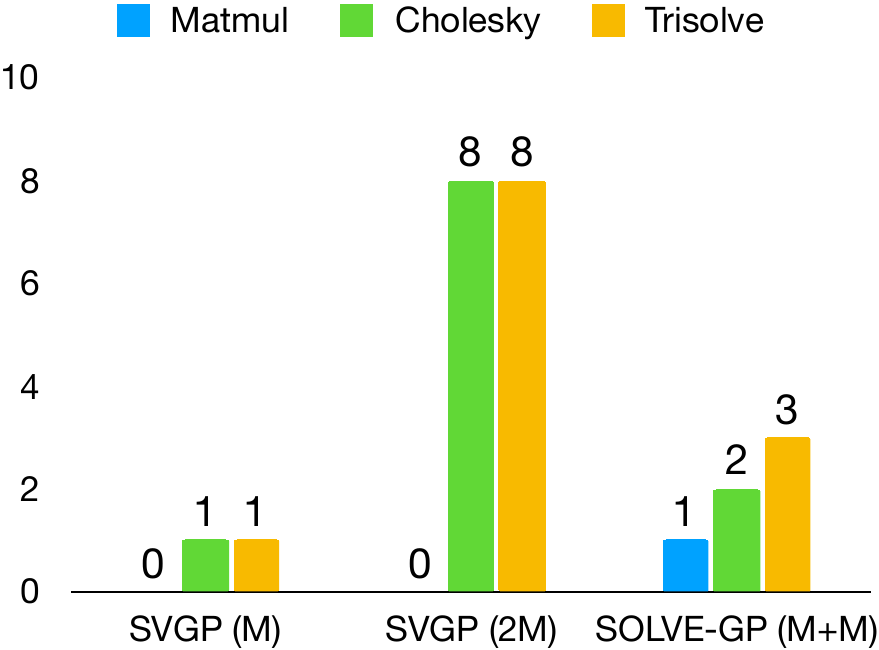}
		\caption{$M \gg N$}
		\label{fig:cost2}
	\end{subfigure}
	\caption{Comparison of computational cost for SVGP and SOLVE-GP. For each method and each type of cubic-cost operation, we plot the factor of increase in cost compared to a single operation on $M\times M$ matrices.}
	\label{fig:cost}
\end{figure}

For comparison purposes, we study two cases of mini-batch training: (i) $N \approx M$ and (ii) $M \gg N$. 
We consider SOLVE-GP with $M_2=M$, which has $2M$ inducing points in total, and then compare to SVGP with $M$ and $2M$ inducing points.  
For each method and each type of operation, we plot the factor of increase in cost compared to a single operation on $M\times M$ matrices.
For instance, when $N\approx M$ (Fig.~\ref{fig:cost1}), SVGP with $M$ inducing points requires solving three triangular matrix equations for $M\times M$ matrices. 
Doubling the number of inducing points in SVGP increases the cost by a factor of $8$, plotted as 24 for SVGP~(2$M$).
In contrast, in SOLVE-GP with $M$ orthogonal inducing points we only need to solve 7 triangular matrix equations for $M\times M$ matrices. 
The comparison under the case of $M\gg N$ is shown in Fig.~\ref{fig:cost2}.
In this case SOLVE-GP additionally introduces one $\mathcal{O}(M^3)$ matrix multiplication operation, but overall the algorithm is still much faster than SVGP (2$M$) given the speed-up in Cholesky decomposition and solving matrix equations.

\begin{table}[tbp]
	\caption{Cubic-cost operations in SOLVE-GP and SVGP, following the implementation in \cref{alg:train}.}
	\label{tab:cost}
	\centering
	{\footnotesize
	\begin{tabular}{ l cc}
		\toprule
		& SVGP & SOLVE-GP \\
		\midrule
		Matrix multiplication & \makecell[r]{$\mathcal{O}(NM^2)\times 1$} & \makecell[r]{$\mathcal{O}(NM^2)\times 1$\\$\mathcal{O}(NM_2^2)\times 1$\\$\mathcal{O}(NMM_2)\times 1$ \\$\mathcal{O}(MM_2^2)\times 1$} \\
		\midrule
		Cholesky & \makecell[r]{$\mathcal{O}(M^3)\times 1$} & \makecell[r]{$\mathcal{O}(M^3)\times 1$ \\$\mathcal{O}(M_2^3)\times 1$} \\
		\midrule
		\makecell[l]{Solving triangular \\matrix equations} & \makecell[r]{$\mathcal{O}(M^3)\times 1$ \\$\mathcal{O}(NM^2)\times 2$} & \makecell[r]{$\mathcal{O}(M^3)\times 1$\\
		$\mathcal{O}(NM^2)\times 2$\\
		$\mathcal{O}(M_2^3)\times 1$\\
		$\mathcal{O}(NM_2^2)\times 2$ \\
		$\mathcal{O}(M_2M^2)\times 1$} \\
		\bottomrule
	\end{tabular}
}
\end{table}

\section{Details of Eq.~\eqref{eq:solve-dgp}}
\label{app:deep-gp}

The variational distribution in Eq.~\eqref{eq:solve-dgp} is defined as:
\begin{equation}
    q(\bu^L, \fperp^L) = \int \prod_{\ell=1}^L\left[p_\perp(\fperp^\ell|\vperp^\ell, \fperp^{\ell-1},\bu^{\ell-1})q(\vperp^\ell)q(\bu^\ell)d\bu^\ell d\vperp^\ell \right] \prod_{\ell=1}^{L-1}d\fperp^\ell.
\end{equation}

\section{Whitening}
\label{app:whiten}

Similar to the practice in SVGP methods, we can apply the ``whitening'' trick~\citep{murray2010slice,hensman2015mcmc} to SOLVE-GP. 
The goal is to improve the optimization of variational approximations by reducing  correlation in the posterior distributions. 
Specifically, we could ``whiten'' $\bu$ by using $\bu' = \Kuu^{-1/2}\bu$, where $\Kuu^{-1/2}$ denotes the Cholesky factor of the prior covariance $\Kuu$. 
Then posterior inference for $\bu$ turns into inference for $\bu'$, which has an isotropic Gaussian prior $\mathcal{N}(\bzero, \bI)$.
Then we parameterize the variational distribution w.r.t.~$\bu'$: $q(\bu') = \mathcal{N}(\bbm_{\bu}, \bS_{\bu})$. 
Whitening $q(\vperp)$ is similar to whitening $q(\bu)$, i.e., we parameterize the variational distribution w.r.t.~$\vperp’ = \Cvv^{-1/2}\vperp$ and set $q(\vperp’) = \mathcal{N}(\bbm_{\bv}, \bS_{\bv})$.
The algorithm can be derived by replacing $\bbm_\bu, \bbm_\bv$ with $\Lu^0\bbm_\bu$, $\Lv^0\bbm_\bv$, and $\bS_\bu, \bS_\bv$ with $\Lu^0\bS_\bu{\Lu^0}^\top, \Lv^0\bS_\bv{\Lv^0}^\top$ in \cref{alg:train} and removing the canceled terms.

\section{Experiment Details}
\label{app:exp-details}

For all experiments, we use kernels with a shared lengthscale across dimensions. All model hyperparameters, including kernel parameters, patch weights in convolutional GP models, and observation variances in regression experiments, 
are optimized jointly with variational parameters using ADAM. 
The variational distributions $q(\bu)$ and $q(\vperp)$ are by default initialized to the prior distributions.
Unless stated otherwise, no ``whitening'' trick~\citep{murray2010slice,hensman2015mcmc} is used for SVGP or SOLVE-GP.

\subsection{1D Regression}

We randomly sample 100 training data points from Snelson's dataset~\citep{snelson2006sparse} as the training data. 
All models use Gaussian RBF kernels and are trained for 10K iterations with learning rate 0.01 and mini-batch size 20. 
The GP kernel is initialized with lengthscale 1 and variance 1. 
The Gaussian likelihood is initialized with variance 0.1.

\subsection{Convolutional GP Models}

All models are trained for 300K iteration with learning rate 0.003 and batch size 64. 
The learning rate is annealed by 0.25 every 50K iterations to ensure convergence.
We use a zero mean function and the robust multi-class classification likelihood~\citep{hernandez2011robust}.
The Gaussian RBF kernels for the patch response GPs in all layers are initialized with lengthscale 5 and variance 5.
We used the TICK kernel~\citep{dutordoir2019translation} for the output layer GP, for which we use a Mat{\'e}rn32 kernel between patch locations with lengthscale initialized to 3.
We initialize the inducing patch locations   to random values in $[0, H]\times[0, W]$, where $[H, W]$ is the shape of the output feature map in patch extraction.

\paragraph{Convolutional GPs} We set patch size to $5\times 5$ and stride to 1. We use the whitening trick in all single-layer experiments for $\bu$ (and $\vperp$) since we find it consistently improves the performance. 
Inducing points are initialized by cluster centers which are generated from running K-means on $M \times 100$ (for SVGP) or $(M + M_2) \times 100$ (for SOLVE-GP) image patches.
The image patches are randomly sampled from 1K images randomly selected from the dataset.

\paragraph{Deep Convolutional GPs} The detailed model configurations are summarized in \Cref{tab:dcgp-config}. No whitening trick is used for multi-layer experiments because we find it hurts performance.
Inducing points in the input layer are initialized in the same way as in the single-layer model. 
In \citet{blomqvist2018deep,dutordoir2019translation}, three-layer models were initialized with the trained values of a two-layer model to avoid getting stuck in bad local minima. 
Here we design an initialization scheme that allows training deeper models without the need of pretraining.
We initialize the inducing points in deeper layers by running K-means on $M \times 100$ (for SVGP) or $(M + M_2) \times 100$ (for SOLVE-GP) image patches which are randomly sampled from the projections of 1K images to these layers. 
The projections are done by using a convolution operation with random filters generated using Glorot uniform~\citep{glorot2010understanding}.
We also note that when implementing the forward sampling for approximating the log-likelihood term, we follow the previous practice~\citep{dutordoir2019translation} to ignore the correlations between outputs of different patches to get faster sampling, which works well in practice.
While it is also possible to take into account the correlation when sampling as this only increases the computation cost by a constant factor, doing this might require multi-GPU training due to the additional memory requirements.

\begin{table*}[htbp] \vskip \baselineskip
\caption{Model configurations of deep convolutional GPs.}
\label{tab:dcgp-config}
\centering
{\footnotesize
\begin{tabular}{ l cc}
\toprule
& 2-layer & 3-layer \\
\midrule
Layer 0 & patch size $5\times 5$, stride 1, out channel 10, & patch size $5\times 5$, stride 1, out channel 10 \\
Layer 1 & patch size $4\times 4$, stride 2 & patch size $4\times 4$, stride 2, out channel 10 \\
Layer 2 & - & patch size $5\times 5$, stride 1 \\
\bottomrule
\end{tabular}
}
\vskip \baselineskip
\end{table*}

\subsection{Regression Benchmarks}

The experiment settings are followed from \citet{wang2019exact}, where we used GPs with Mat{\'e}rn32 kernels and 80\% / 20\% training / test splits. 
A 20\% subset of the training set is used for validation. 
We repeat each experiment 5 times with random splits and report the mean and standard error of the performance metrics. 
For all datasets we train for 100 epochs with learning rate 0.01 and mini-batch size 1024.

\section{Additional Results}
\label{app:exp}

\subsection{Regression Benchmarks} 

Due to space limitations in the main text, we include the Root Mean Squared Error (RMSE) on test data in \Cref{tab:rmse}.
The results on Elevators and Bike are shown in \Cref{tab:reg-ext}.

\begin{table*}[htbp] \vskip \baselineskip
\caption{Test RMSE values of regression datasets. The numbers in parentheses are standard errors. Best mean values are highlighted, and asterisks indicate statistical significance.} %
\label{tab:rmse}
\makebox[\textwidth][c]{
\resizebox{1.02\textwidth}{!}{
\setlength\tabcolsep{2pt}
\begin{tabular}{ l l cc cc cc cc}
\toprule
& &  Kin40k & Protein &  \footnotesize KeggDirected &  KEGGU & 3dRoad & Song & Buzz & \footnotesize HouseElectric \\
\cmidrule{3-10}
& $N$ & 25,600 & 29,267 & 31,248 & 40,708 & 278,319 & 329,820 & 373,280 & 1,311,539  \\
& $d$ & 8 & 9 & 20 & 27 & 3 & 90 & 77 & 9  \\
\midrule
\multirow{2}{*}{SVGP}
& $1024$  
& 0.193(0.001) & 0.630(0.004) & 0.098(0.003) & {\bf 0.123}(0.001) & 0.482(0.001) & 0.797(0.001) & 0.263(0.001) & 0.063(0.000) \\
& $1536$  
& 0.182(0.001) & 0.621(0.004) & 0.098(0.002) & {\bf 0.123}(0.001) & 0.470(0.001) & 0.797(0.001) & 0.263(0.001) & 0.063(0.000) \\
\midrule
\multirow{2}{*}{\shortstack{ODVGP}} & $1024+1024$ 
& 0.183(0.001) & 0.625(0.004) & 0.176(0.012) & 0.156(0.004) & 0.467(0.001) & 0.797(0.001) & 0.263(0.001) & 0.062(0.000) \\
& $1024+8096$  
& 0.180(0.001) & 0.618(0.004) & 0.157(0.009) & 0.157(0.004) & {\bf 0.462}(0.002) & 0.797(0.001) & 0.263(0.001) & 0.062(0.000) \\
\midrule
\footnotesize \multirow{1}{*}{SOLVE-GP} &$1024 + 1024$ 
& *{\bf 0.172}(0.001) & 0.618(0.004) & {\bf 0.095}(0.002) & {\bf 0.123}(0.001) & 0.464(0.001) & {\bf 0.796}(0.001) & {\bf 0.261}(0.001) & *{\bf 0.061}(0.000) \\
\midrule
\multirow{1}{*}{\shortstack{SVGP}}
& $2048$
& 0.177(0.001) & {\bf 0.615}(0.004) & 0.100(0.003) & 0.124(0.001) & 0.467(0.001) & {\bf 0.796}(0.001) & 0.263(0.000) & 0.063(0.000) \\
\bottomrule
\end{tabular}
}}
\vskip \baselineskip
\end{table*}

\begin{table*}[htbp]
\caption{Regression results on Elevators and Bike. Best mean values are highlighted.}
\label{tab:reg-ext}
\centering
 \resizebox{0.7\textwidth}{!}{
\begin{tabular}{ ll cc cc}
\toprule
& & \multicolumn{2}{c}{\makecell{Elevators\\$N=10,623,\; d=18$}} & \multicolumn{2}{c}{\makecell{Bike\\$N=11,122,\; d=17$}} \\
 \cmidrule(l){3-4} \cmidrule(l){5-6}
& & Test LL & RMSE & Test LL & RMSE \\
\midrule
\multirow{2}{*}{SVGP}
& $1024$  
& -0.516(0.006) & 0.398(0.004) & -0.218(0.006) & 0.283(0.003) \\
& $1536$  
& -0.511(0.007) & 0.396(0.004) & -0.203(0.006) & 0.279(0.003) \\
\midrule
\multirow{2}{*}{\shortstack{ODVGP}} & $1024+1024$ 
& -0.518(0.006) & 0.397(0.004) & -0.191(0.006) & 0.272(0.003) \\
& $1024+8096$  
& -0.523(0.006) & 0.399(0.004) & {\bf -0.186}(0.006) & {\bf 0.270}(0.003) \\
\midrule
\multirow{1}{*}{SOLVE-GP} &$1024 + 1024$ 
& -0.509(0.007) & {\bf 0.395}(0.004) & -0.189(0.006) & 0.272(0.003) \\
\midrule
\multirow{1}{*}{\shortstack{SVGP}}
& $2048$ 
& {\bf -0.507}(0.007) & {\bf 0.395}(0.004) & -0.193(0.006) & 0.276(0.003) \\
\bottomrule
\end{tabular}
 }
\end{table*}

\subsection{Convolutional GP Models}
We include here the full tables for CIFAR-10 classification, where we also report the accuracies and predictive log-likelihoods on the training data. \Cref{tab:app-conv} contains the results by convolutional GPs. \Cref{tab:app-deep-conv2} and \Cref{tab:app-deep-conv3} include the results of 2/3-layer deep convolutional GPs.

\begin{table*}[htbp]
\caption{Convolutional GPs for CIFAR-10 classification.}
\label{tab:app-conv}
\centering
{\footnotesize
\begin{tabular}{ l l cc cc c}
\toprule
& & Train Acc & Train LL & Test Acc & Test LL & Time \\
\midrule
\multirow{2}{*}{SVGP}
& $1000$ & 77.81\% & -1.36 & 66.07\% & -1.59 & 0.241 s/iter \\
& $1600$ & 78.44\% & -1.26 & 67.18\% & -1.54 & 0.380 s/iter \\
\multirow{1}{*}{SOLVE-GP} & $1000+1000$ & 79.32\% & -1.20 & \bf 68.19\% &  \bf -1.51 & 0.370 s/iter \\
\midrule
\multirow{1}{*}{\shortstack{SVGP}}
& $2000$ & 79.46\% & -1.22 & 68.06\% & \bf -1.48 & 0.474 s/iter \\
\bottomrule
\end{tabular}
}
\end{table*}

\begin{table*}[htbp]
\caption{2-layer deep convolutional GPs for CIFAR-10 classification. }
\label{tab:app-deep-conv2}
\centering
{\footnotesize
\begin{tabular}{ l l cc cc c}
\toprule
& Inducing Points & Train Acc & Train LL & Test Acc & Test LL & Time \\
\midrule
SVGP & $384,1\text{K}$ & 84.86\% & -0.82 & 76.35\% & -1.04 & 0.392 s/iter  \\
SOLVE-GP & $384+384,1\text{K}+1\text{K}$ & 87.59\% & -0.72 & \bf 77.80\% & \bf -0.98 & 0.657 s/iter \\
\midrule
SVGP & $768,2\text{K}$ & 87.25\% & -0.74 & 77.46\% & \bf -0.98 & 1.104 s/iter  \\
\bottomrule
\end{tabular}
}
\end{table*}

\begin{table*}[htbp]
\caption{3-layer deep convolutional GPs for CIFAR-10 classification. }
\label{tab:app-deep-conv3}
\centering
{\footnotesize
\begin{tabular}{ l l cc cc c}
\toprule
& Inducing Points & Train Acc & Train LL & Test Acc & Test LL & Time \\
\midrule
SVGP & $384,384,1\text{K}$ & 87.70\% & -0.67 & 78.76\% & -0.88 &  0.418 s/iter \\
SOLVE-GP & $(384+384)\times 2,1\text{K}+1\text{K}$ & 89.88\% & -0.57 & \bf 80.30\% & \bf -0.79 & 0.752 s/iter \\
\midrule
SVGP & $768,768,2\text{K}$ & 90.01\% & -0.58 & \bf 80.33\% & -0.82 & 1.246 s/iter \\
\bottomrule
\end{tabular}
}
\end{table*}

\end{document}